\documentclass[twoside,11pt]{article}
\usepackage{algorithm}
\usepackage{graphicx}
\usepackage{siunitx}
\usepackage{algpseudocode}
\def\G3{G$^3$}
\usepackage[tight]{subfigure}
\usepackage[percent]{overpic}
\usepackage{flushend}
\usepackage{amsmath}
\usepackage{setspace}
\usepackage{booktabs}
\usepackage{amsmath}
\usepackage{amssymb}
\usepackage{hyperref}
\usepackage{changes}
\usepackage{qtree}
\usepackage{com.braju.graphicalmodels}
\catcode`\@=11%
\newcommand{\argmax}[1]{\underset{#1}{\operatorname{argmax}}\;}

\usepackage{color}
\definechangesauthor[color=red]{NR}
\definechangesauthor[color=blue]{ST}
\definechangesauthor[color=green]{TK}

\usepackage{todonotes}
\makeatletter
\setremarkmarkup{\todo[color=Changes@Color#1!20,size=\scriptsize]{#1: #2}}
\makeatother

\usepackage{natbib}

\begin{document}

\title{\vspace{-1in}Generalized Grounding Graphs: A Probabilistic Framework for
  Understanding Grounded Commands} 

\author{Thomas Kollar$^1$, Stefanie Tellex$^1$, Matthew R.\ Walter,\\
  Albert Huang, Abraham Bachrach, Sachi Hemachandra, Emma Brunskill, \\
Ashis Banerjee, Deb Roy, Seth Teller, Nicholas Roy}


\maketitle

\begin{abstract}
%

Many task domains require robots to interpret and act upon natural
language commands which are given by people and which refer to the
robot's physical surroundings. Such interpretation is known variously
as the symbol grounding problem \citep{harnad90}, grounded semantics
\citep{feldman1996l_0} and grounded language acquisition
\citep{nenov1993perceptually, nenov1994perceptually}.  This problem is
challenging because people employ diverse vocabulary and grammar, and
because robots have substantial uncertainty about the nature and
contents of their surroundings, making it difficult to associate the
constitutive language elements (principally noun phrases and spatial
relations) of the command text to elements of those surroundings.
Symbolic models capture linguistic structure but have not scaled
successfully to handle the diverse language produced by untrained
users.  Existing statistical approaches can better handle diversity,
but have not to date modeled complex linguistic structure, limiting
achievable accuracy.  Recent hybrid approaches have addressed
limitations in scaling and complexity, but have not effectively
associated linguistic and perceptual features.  Our framework, called
Generalized Grounding Graphs (\G3), addresses these issues by defining
a probabilistic graphical model dynamically according to the
linguistic parse structure of a natural language command.  This
approach scales effectively, handles linguistic diversity, and enables
the system to associate parts of a command with the specific objects,
places, and events in the external world to which they refer.  We show
that robots can learn word meanings and use those learned meanings to
robustly follow natural language commands produced by untrained users.
We demonstrate our approach for both mobility commands (e.g.\ route
directions like ``Go down the hallway through the door'') and mobile
manipulation commands (e.g.\ physical directives like ``Pick up the
pallet on the truck'') involving a variety of semi-autonomous robotic
platforms, including a wheelchair, a micro-air vehicle, a forklift,
and the Willow Garage PR2.
\addtocounter{footnote}{1} 
\footnotetext{The first two authors contributed equally to this paper.}
\end{abstract}


\section{Introduction}
\label{chapter:introduction}\label{sec:grounding}

\begin{figure}
\begin{center}
\subfigure[Robot forklift.\label{fig:exampleForklift}]{\fbox{
\parbox{0.24\linewidth}{\tiny
\includegraphics[width=1\linewidth]{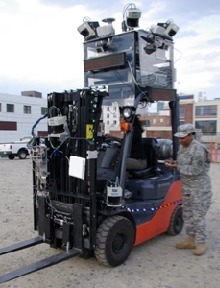}
Place the pallet of tires on the left side of the trailer.\\\\
Arrange tire pallet to the truck.\\\\
Please lift the set of six tires up and set them on the trailer, to the right of the set of tires already on it.\\\\
Load the skid right next to the other skid of tires on the trailer.\\\\
Place a second pallet of tires on the trailer.\\\\
Put the tire pallet on the trailer.\\\\
Place the pallet of tires on the right side of the truck.
\\\\\\\\
}}}~%
\subfigure[Robot wheelchair.\label{fig:exampleWheelchair}]{\fbox{\parbox{0.24\linewidth}{\tiny 
  \includegraphics[width=1\linewidth]{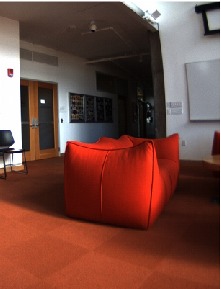}
With your back to the windows, walk straight through the door near the elevators.  Continue to walk straight, going through one door until you come to an intersection just past a whiteboard.  Turn left, turn right, and enter the second door on your right (sign says ``Administrative Assistant'').\\
\\
Go through the set of double doors by the red couches.  
Continue straight down the corridor, pass through two sets of wooden double doors, and enter through the gray door facing you with a phone next to it.  Take a left through the blue double doors, and follow the corridor.  The room is the second door on the left after you have rounded the corner.
}}}~%
\subfigure[PR2.\label{fig:examplePR2}]{\fbox{\parbox{0.24\linewidth}{\tiny 
  \includegraphics[width=1\linewidth]{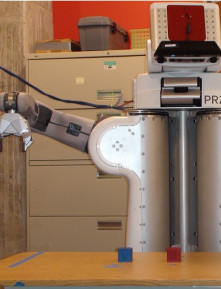}
Pick up the red block.\\
\\\\
Pick up the blue block.
\\\\
Put the red block near the blue block.
\\\\
Put the blue block behind the red block. 
\\\\
Pick up the blue box and keep it on the right side. 
\\\\
Pick up the blue block and place it to the far right of the orange block.
\\\\\\\\\\\\
\\\\\\
}}}~%
\subfigure[Micro-air vehicle.\label{fig:exampleMav}]{\fbox{\parbox{0.24\linewidth}{\tiny 
  \includegraphics[width=1\linewidth]{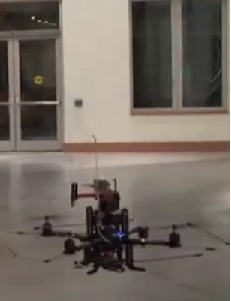}
 Go forward until you
    are able to make a left.  Then move ahead until you reach the
    opposite wall, then make a right.  Go straight, past one staircase
    and to the next staircase.  At the top of this staircase you will
    find a fire alarm on your left at approximately 7ft up. 
\\
\\
Turn right and go ahead through the gap between two walls to a large
white column.  Turn right again and move until you are the base of the
staircase to your right.  Ahead, suspended from the ceiling is a large
white butterfly sculpture. ({\em sic})
\\
\\
Go forward ten feet and turn right. Go down the hall and over the
stairs. Look straight ahead at the wall at the end of the walkway.
}}}

\end{center}
\caption{Four domains studied in this paper, along with sample
  commands from the evaluation corpora used for each domain.
\label{fig:exampleDomains}}
\end{figure}

To be useful teammates to human partners, robots must be able to
robustly follow spoken instructions. For example, a human supervisor
might tell an autonomous forklift, ``Put the tire pallet on the
truck,'' or the occupant of a wheelchair equipped with a robotic arm
might say, ``Get me the book from the coffee table.''  Understanding
such commands is challenging for a robot because they involve verbs
(``Put''), noun phrases (``the tire pallet''), and prepositional
phrases (``on the truck''), each of which must be grounded to aspects
of the world and which may be composed in many different ways.
Figure~\ref{fig:exampleDomains} shows some of the wide variety of
human-generated commands that might be given to different robots in
different situations.

Traditional approaches, starting with \citet{winograd71}, have
manually created symbol systems that map between language and the
external world, connecting each linguistic term onto a space of
pre-specified actions and environmental
features~\citep{hsiao2003coupling, roy2003crb, bugmann04, roy05,
  macmahon06, kress-gazit08, dzifcak09}. This class of systems takes
advantage of the structure of language, but usually does not involve
learning and has a fixed action space, limiting the ability of the
resulting system to robustly understand language produced by untrained
users.

Statistical approaches address these limitations by using data-driven
training to learn robust models of word meanings~\citep{
  shimizu2009learning, regier92, branavan09, branavan12, vogel10}.
However, these approaches use a fixed and flat sequential structure
that does not capture the argument structure of language
(e.g., it does not allow for variable arguments or nested clauses).
At training time, a system that assumes a flat structure sees the
entire phrase ``the pallet beside the truck'' and has no way to
separate the meanings of relations like ``beside'' from objects such
as ``the truck.''  Furthermore, a flat structure ignores the argument
structure of verbs.  For example, the command ``put the box pallet on
the ground beside the truck,'' has two arguments (``the box pallet''
and ``on the ground beside the truck''), both of which must be
isolated in order to infer the appropriate meaning of the verb ``put''
in this instance.  To infer the meaning of unconstrained natural
language commands, it is critical for the model to reason over these
compositional and hierarchical linguistic structures at both learning
and inference time.

Existing approaches that combine statistical and symbolic approaches
assume access to a predefined lexicon of symbols that they then learn
to map to natural language~\citep{artzi13, chen11, matuszek12,
  ge:conll05, liang11, arumugam17, misra16}.  These symbols must be
provided by the designer to the system in advance.  Newer
approaches~\citep{MisraLA17} use deep learning in an MDP setting to
map between actions and visual features, but do not decompose the
language command according to the parse structure, which makes it
harder to reason about what parts of the command were and were not
understood by the system.


To address these issues, we present a framework called Generalized
Grounding Graphs (\G3).  A {\em grounding graph} is a probabilistic
graphical model defined dynamically according to the compositional and
hierarchical structure of a natural language command. The model
predicts physical interpretations or {\em groundings} for linguistic
constituents.  Groundings are specific physical concepts that are
referred to by the language and can be objects (e.g., a truck or a
door), places (e.g., a particular location in the world), paths (e.g.,
a trajectory through the environment), or actions / events (e.g., a
sequence of actions taken by the robot).  The system is trained in a
supervised way, using a corpus of language paired with groundings,
enabling it to learn probabilistic predicates that map between
language and groundings from lower-level features, without
prespecifying the predicates in advance.  At inference time, the
system is given a natural language command and infers the most
probable set of groundings in the external world.  For example, for a
command such as ``Put the tire pallet on the truck,'' the system
infers that the noun phrase ``the tire pallet'' maps to a specific
pallet in the robot's representation of nearby objects, and the
prepositional phrase ``on the truck'' maps to a particular location in
the environment.  For the entire sentence, the robot infers a specific
sequence of actions that it should execute.

We evaluate \G3 on four robotic domains: a robotic forklift, the PR2
mobile manipulator, a robotic wheelchair, and a robotic micro-air
vehicle (MAV).  The forklift domain, shown in
Figure~\ref{fig:exampleForklift}, considers mobile manipulation
commands, such as ``Pick up the tire pallet.''  This domain shows that
our approach is able to understand hierarchical and compositional
commands, learning verbs, such as ``put'' and ``take,'' as well as
spatial relations, such as ``on'' and ``to.''  The wheelchair and MAV
domains demonstrate the ability to interpret longer route direction
commands for ground and aerial
robots. Figures~\ref{fig:exampleWheelchair} and \ref{fig:exampleMav}
shows samples of these longer commands from the route direction
domain.  Performing inference in the full hierarchical model is
computationally expensive and becomes intractable when understanding
these longer route directions.  We present an approximation of the
full hierarchical structure that is able to successfully follow many
route directions to within 10 meters of the destination, along with
real-world demonstrations of our approach on a micro-air vehicle.

This article provides the definitive statement of the \G3 framework,
unifying our previous work in this area~\citep{kollar10, huang10,
  kollar2010grounding, tellex2011}.  It precisely specifies the
technical details of our approach, and provides corpus-based
evaluations and robotic demonstrations in several domains.
\citet{tellex11a} gives a more reflective overview of the framework,
focusing on learned word meanings, but that work does not include our
results on indoor mobility domains, and does not present all domains
in a unified technical framework.  In our more recent work, we have
extended the \G3 framework in several ways.  \citet{tellex13} and
\citet{deits13} showed how to use \G3 to ask questions based on
entropy when the robot is ``confused,'' i.e.\ when its uncertainty is
high.  The \G3 framework can also be used by the robot to generate a
request for help, by inverting the semantics model~\citep{knepper13}.
\citet{bollini12} applied the framework to the cooking domain,
enabling a robot to follow natural language recipes.  \citet{tellex13}
described how to train the \G3 framework with less
supervision. \citet{kollar13a} and \citet{kollar13b} introduced a
compositional parser based on a combinatory categorial grammar (CCG);
the parser can be trained jointly using limited supervision.
Expanding beyond understanding, \cite{kollar13c} used dialog to learn
new symbols and referring expressions for physical locations, enabling
robots to execute commands to find and fetch objects.  Similar to the
local version of the \G3 model, \cite{kollar13d} described an approach
for learning a policy that follows directions explicitly in unknown
environments. Most recently, \citet{howard14} and \citet{paul16}
described hierarchical extensions that enable model inference over
learned abstractions.

\section{Related Work}
\label{chapter:related_work}\label{sec:sensorimotor}

The problem of robust natural language understanding has been studied
since the earliest days of artificial intelligence.  Beginning with
SHRDLU~\citep{winograd71}, many systems have exploited the
compositional structure of language to generate symbolic
representations of the natural language input, for example with
approaches based on formal logic~\citep{kress-gazit08, dzifcak09} or
symbol grounding~\citep{skubic_spatial_2004}.  Notably,
\citet{macmahon06} developed a symbol-based system for following route
directions through simulated environments, and
\citet{hsiao_object_2008} created a system for enabling a humanoid
robot to follow natural language commands involving manipulation of
table-top objects.  These systems exploit the structure of language,
but usually do not involve learning, and have a fixed action space.
Our work, in contrast, defines a probabilistic graphical model
according to the structure of the natural language command, inducing a
probability distribution over the mapping between words and groundings
in the external world.  This factored structure enables the system to
understand novel commands never seen during training, by
compositionally combining learned word meanings according to the
factorization induced by the command.

Another approach is to associate language with different aspects of
the environment, typically by employing statistical methods to learn
the mapping.  \citet{harnad90} called the problem of mapping between
words in the language and aspects of the external world the ``symbol
grounding problem.'' Some systems learn the meaning of words directly
in the sensorimotor space (e.g., joint angles and images) of the
robot~\citep{roy02, sugita2005learning, modayil_2007, marocco_2010}.
By treating linguistic terms as a sensory input, these systems must
learn directly from complex features extracted by perceptual modules,
limiting the set of commands that they can robustly understand, and
their ability to handle complex syntactic structures.  Other
computational approaches use more complex syntactic structures but can
handle just a few words, such as ``above,''~\citep{regier92} or
``near''~\citep{carlson05}; \citet{tellex10a} describe models for just
six spatial prepositions applied to the problem of video retrieval.
Our work, in contrast, learns grounded word meanings for the much
wider variety of words that appear in a training set.


A different approach is to apply semantic parsing, which maps natural
language sentences to symbolic semantic representations.  Early work
in semantic parsing uses labeled data that consists of sentences
paired with their logical form, beginning with \citet{thompson03} on
the GeoQuery dataset, which consists of natural language queries about
geography.  More recent work that considers this dataset and others
provided improved training methods~\citep{zettlemoyer05, wong07,
  piantadosi08, kwiatkowski10}.  Some of these approaches have been
applied to command understanding for robotics~\citep{chen11,
  matuszek12, artzi13}.  However these approaches require the
designers to provide relational predicates to the system that map
between natural language and world knowledge, such as $\textsc{to}$ or
$\textsc{near}$.  Our approach, in contrast, learns predicates using
lower-level features.  \citet{shimizu_2006} and
\citet{shimizu2009learning} train a CRF to map from language to a
small fixed command set for following route directions; grounding
occurs by applying these commands to a backend executor. Newer
approaches learn symbolic word meanings with less supervision;
\citet{poon09} presented an approach to unsupervised semantic parsing,
while other approaches use an external reward signal~\citep{liang11,
  clarke10}.  Our work, in contrast, requires more supervision, but
learns grounded meaning representations in terms of perceptual
features rather than symbols.  \citet{liang06} created an end-to-end
discriminative approach to machine translation that introduced a
correspondence structure between the input and output translations.
As in their approach, we optimize over all possible groundings within
a correspondence structure that allows us to introduce features that
measure the mapping between words and groundings, measuring the
faithfulness of particular groundings in the environment as they
correspond to words in the language.


\citet{cohen95} report that speech interfaces are useful when the
hands and eyes are otherwise engaged and when there is limited
availability of keyboards or screens.  Robots that operate in
unstructured, real-world environments fit these scenarios perfectly.
Despite these characteristics of human-robot interaction problem,
there is no consensus that human-robot interfaces should be built
around natural language, due to the challenges in building dialog
interfaces in changing social contexts~\citep{fong03, severinson03}.
The aim of our work is to develop robust natural language systems so
that robots can interact with people flexibly using language.

Other recent work built upon semantic parsing by explicitly
considering perception at the same time as
parsing~\citep{matuszek12joint, kollar13a,
  kollar13b,andreas2016learning}.  These approaches used limited
supervision to learn models that connect language to sets of objects,
attributes and relations.  Other work used attention-based neural
networks to answer questions about
images~\cite{saenko15,smola15,zitnick2016measuring}, which do not
generally use an intermediate parse.  Although \G3 uses full
supervision to train its semantic parser and grounding classifiers, it
is able to additionally learn to understand verb phrases such as
``pick up'' and ``put,'' which are challenging because such learning
involves parameter estimation and inference over state sequences.

Newer approaches use learning in an MDP setting to map between actions
and visual features.  Some existing approaches learn policies to map
language to the correct action~\citep{branavan09, vogel10}.  These
approaches can learn from weaker supervision but do not explicitly
learn word meanings or probabilistic predicates as in our approach.
\citet{misra17} learns to map between actions in a predefined space
and visual observations in an integrated system, but does not
decompose word meanings into separately trained factors.  Separately
trained models allow us to decompose word meanings into separate
factors and recombine them later to understand novel commands, as well
as learn spatially relevant word meanings that can be applied later in
different settings.  \citet{janner17} created a system for spatial
reasoning for a mobile robot that uses deep reinforcement learning.
This approach automatically learns end poses for a mobile robot,
including relative expressions like ``above the westernmost rock.''
However it only handles movement commands and does not do
manipulation.  Additionally, the approach does not decompose into
separate factors, which makes it harder to ask targeted questions.
\citet{andreas15} created a system that explicitly models low-level
compositional structure for instruction following.  They use the same
technical term, {\em grounding graph} to describe the perceptual
representation of the world model, rather than the grounding structure
to map between language and the external world.  By using graph
semantics, the work created a map between non-binary features and
learned concepts such as ``you are on top of the hill.''

\section{Generalized Grounding Graphs}
\label{sec:g3}

\begin{table}
\small
\begin{tabular}{cc}
\toprule
Variable   & Description\\
\midrule
$\Gamma : \gamma_1 \dots \gamma_N$ &\parbox{0.8\linewidth}{Grounding; can be an object, place, path or event in the world. Each $\gamma_n$ consists of a tuple $(g, t, p)$.}\\
$\Lambda : \lambda_1 \dots \lambda_M$&\parbox{0.8\linewidth}{Language command.  Each linguistic constituent $\lambda_m$ refers to a subset of grounding variables based on the parse structure of the command.}\\
$\Phi : \phi_1 \dots \phi_M$    &\parbox{0.8\linewidth}{Correspondence variable; Each there is a $\phi_m \in \left\{ 0,1\right\}$  for each $\lambda_m$ and a subset of the grounding variables $\gamma_n$.}\\ 
$\Psi: \psi_1 \dots \psi_M$     & \parbox{0.8\linewidth}{Factors defined according to the parse structure of the command.}\\ 
$M$        &\parbox{0.8\linewidth}{Environment model; the robot's model of the external world, acquired from its sensors.}\\ 
$Z$        &\parbox{0.8\linewidth}{Normalization constant.}\\ 
$g$        &\parbox{0.8\linewidth}{A three-dimensional shape of an object grounding $\gamma_m$, such as a
  room, or the vicinity of an object.  It is expressed as a set of points that define
  a polygon $(x_1, y_1), \dots, (x_N, y_N)$ together with a height
  $z$ (e.g., as a vertical prism).  }\\ 
$p \in \mathbb{R}^{T \times 7}$        &\parbox{0.8\linewidth}{a sequence of $T$ points.
  Each point is a pose for the region, $g$.  It consists of a tuple
  $(\tau, x, y, z, roll, pitch, yaw)$ representing the location and
  orientation of $g$ at time $\tau$ (with location interpolated
  linearly across time).  The combination of the shape, $g$, and the
  trajectory, $T$, define a three-dimensional region in the
  environment corresponding to the object at each time step.}\\ 
$t$        &\parbox{0.8\linewidth}{A set of pre-defined textual tags $\{tag_1, \dots,
  tag_M\}$ that are the output of perceptual classifiers, such as
  object recognizers or scene classifiers.}\\ 
$s_j$        &\parbox{0.8\linewidth}{Feature functions.}\\ 
$\theta_j$        &\parbox{0.8\linewidth}{Feature weights.}\\ 
\bottomrule
\end{tabular}
\caption{Table of variables.\label{tab:variables}}
\end{table}

To understand natural language commands, a robot must be able to map
between the linguistic elements of a command, such as ``Pick up the
tire pallet,'' and the corresponding aspects of the external world.
Each constituent phrase in the command refers to a particular object,
place, or action that the robot should execute in the environment.  We
refer to the object, place, path, or event as the {\em grounding} for
a linguistic constituent.  The aim of language understanding is to
find the set of most probable groundings $\Gamma$ given a parsed
natural language command $\Lambda$ and the robot's model of the
environment, $M$.  
\begin{align}
\argmax {\Gamma} p(\Gamma | \Lambda, M).
\label{eq:g3_argmax}
\end{align}

The environment model $M$ consists of the robot's location along with
the locations, structure and appearance of objects and places in the
external world.  It defines a space of possible values for the set of
groundings $\gamma_i\in\Gamma$.  A robot computes the environment
model using sensor input.  An object such as a pallet is represented
as a three-dimensional geometric shape, along with a set of symbolic
tags that might be produced by an object classifier, such as
``pallet'' or ``truck.''  A place grounding represents a particular
location in the map, corresponding to a phrase such as ``on the
truck.''  Path and action groundings are constrained by the
environment, but are not fully defined by it, since the continuous
state space of the environment defines an infinite space of paths for
the robot and objects.  As such, the environment model defines the
starting location for the robot and the objects in the environment,
from which paths of the robot and objects can be derived.  A path
grounding is a trajectory of points over time, corresponding to a
phrase such as ``to the elevators.''  An action or event grounding
consists of the robot's trajectory over time as it performs some
action (e.g., as it picks up a pallet).\footnote{In general, an event
  could consist of any state change in the environment, for example
  the generation of an appropriate sound for the command, ``Speak your
  name.''  In this paper, we focus on events that can be represented
  geometrically. } The groundings $\Gamma$ correspond to $n$ items
$\gamma_1 \dots \gamma_n$, where each $\gamma_i$ is a tuple, $(g, t,
p)$; $g$ is the three dimensional shape of the object, $t$ is a
trajectory through space, and $p$ is a set of symbolic perceptual
tags.  These variables are defined more formally in
Table~\ref{tab:variables}.



Since learning the joint distribution over commands and groundings is
intractable, we must factor Equation~\ref{eq:g3_argmax} into simpler
components.  Natural language has a well-known compositional,
hierarchical argument structure~\citep{jackendoff83}, dividing a
sentence into a parse tree where each node is one of the {\em
  linguistic constituents} $\lambda_i$ ~\citep{akmajian10_chapter5}
produced by the parser.  Each linguistic constituent has corresponding
grounding variables $\left\{ \gamma_i , \dots , \gamma_{i+k}\right\}
\in \Gamma$ that are automatically instantiated based on the structure
of the parse and is scored using a factor $\psi_m$; each factor
$\psi_m$ defines the score of a \emph{probabilistic predicate} for the
linguistic constituent $\lambda_i$, and factors as follows:

\begin{align}
p(\Gamma | \Lambda, M) = \frac{1}{Z} \prod_m \psi_m(\lambda_m, \gamma_{m_1} \dots \gamma_{m_k}).
\label{eq:factored_problem_statement_1}
\end{align}

\begin{figure*}[t]
  \centering 
  \subfigure[\label{figure:parse}]{
    \begin{minipage}[b]{0.43\linewidth}
      \underline{Language $\Lambda$} \\ 
                {\tiny
                  ``put the pallet on the truck'' \\ 
                }\\ \\
      \underline{Linguistic constituents $\lambda_i$} \\ 
      \tiny\tt 
      (ROOT (S (VP (VB Put) \newline
      \hspace*{0.31in}~~~~~~(NP (DT the) (NN pallet))\newline
      \hspace*{0.31in}~~~~~~(PP (IN on)\newline
      \hspace*{0.31in}~~~~~~~~~~(NP (DT the) (NN truck))))\newline
      \hspace*{0.01in}~(. .)))
      \newline
    \end{minipage} 
  }\subfigure[\label{figure:environment:M}]{
    \begin{minipage}[b]{0.28\linewidth} 
      \underline{Environment $M$} \\ 
                {\parbox{0.95\linewidth}{\centering\includegraphics[width=0.95\linewidth]{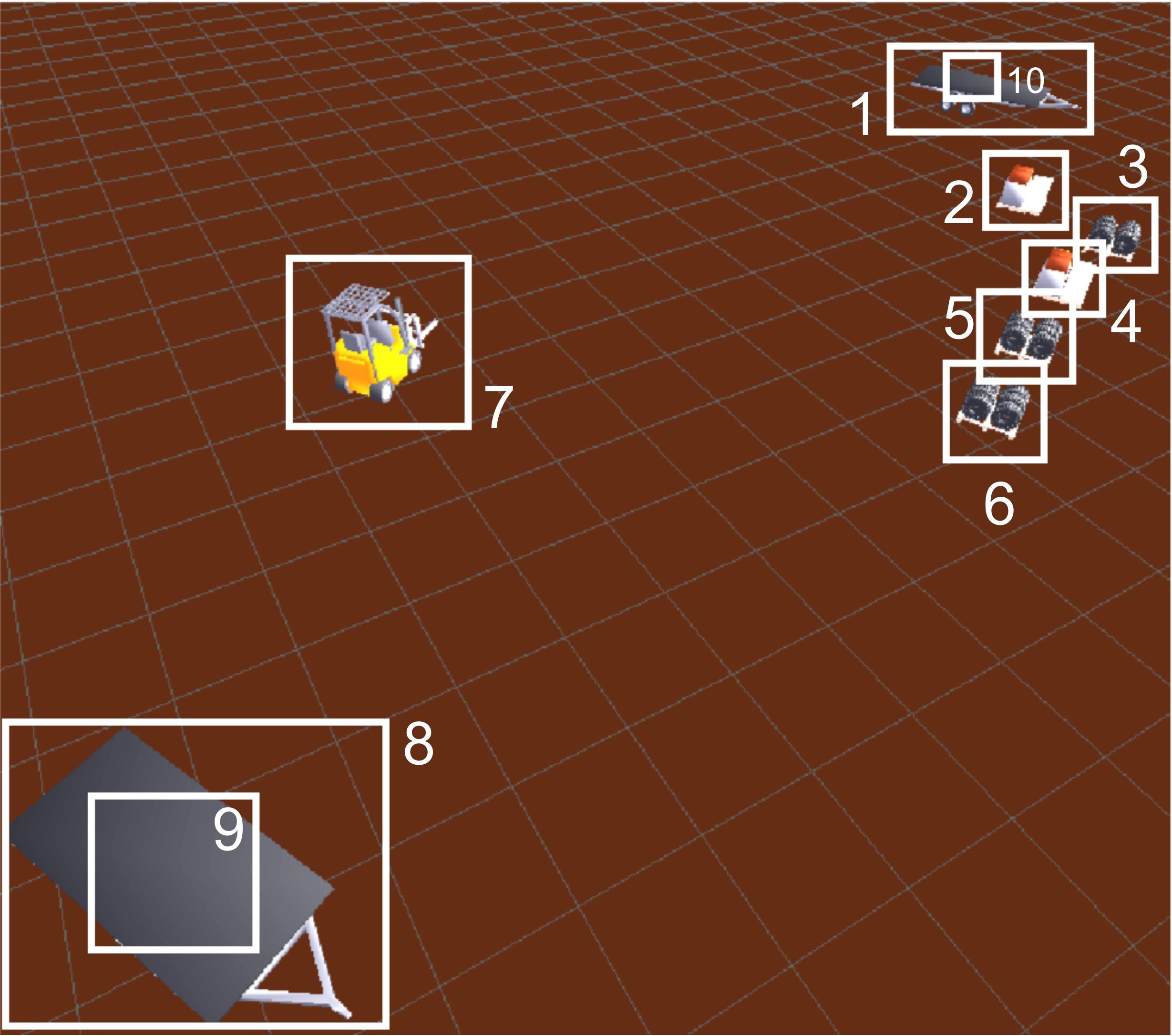}}}\\
    \end{minipage}
  }\subfigure[\label{figure:path_groundings}]{
    \begin{minipage}[b]{0.28\linewidth}
    \underline{Path Groundings} \\ 
              {\parbox{0.95\linewidth}{\centering\includegraphics[width=0.95\linewidth]{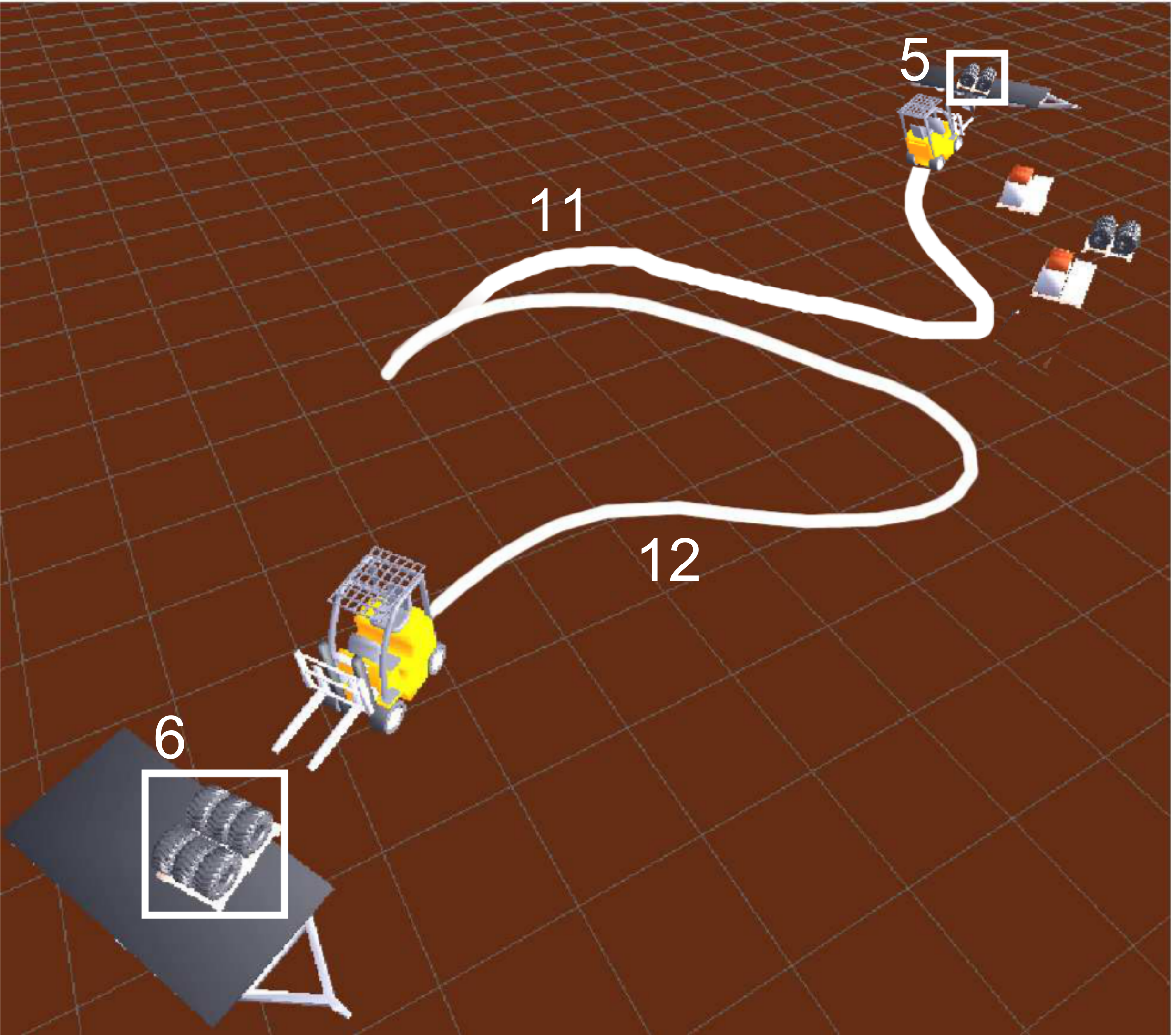}}}\\
    \end{minipage}
  }\\
\subfigure[\label{figure:grounding_graph_ex}]{
    \begin{minipage}[b]{0.5\linewidth}   \underline{Grounding graph} \\
      \includegraphics[width=0.9\linewidth]{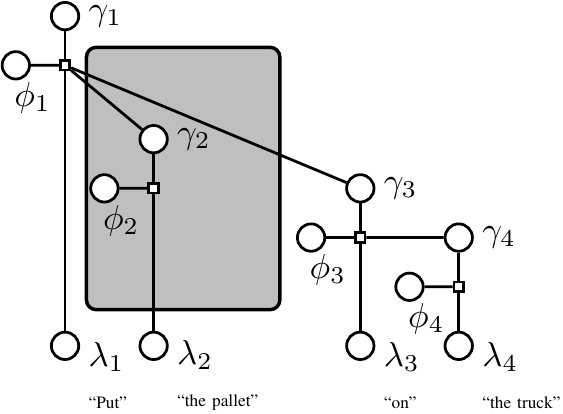}\\
    \end{minipage}    
  } \subfigure[\label{figure:overview:parsing}]{
    \begin{minipage}[b]{0.45\linewidth} \underline{Probabilistic predicates} \\
      {\footnotesize
        \begin{tabular}{rl}
          \toprule
          $\lambda_i$ & $(\gamma_{m_1}\dots \gamma_{m_k})$\\
          \midrule 
          \multicolumn{2}{l}{\bf{entity}}\\
          \quad ``the pallet'' & $\gamma_2 \in \{2,3,4,5,6\}$ \\ 
          \quad ``the truck'' & $\gamma_4 \in \{1,8\}$\\
          \multicolumn{2}{l}{\bf{relation}}\\
          ``on'' & $\left(\gamma_3, \gamma_4\right) \in$\\
          & \qquad $\{(9,8),(10,1)\}$ \\ 
          ``put'' & $\left(\gamma_1, \gamma_2, \gamma_3\right) \in$ \\ 
          & \qquad$\{(11,5,10),(12,6,9)\}$ \\
          \bottomrule 
        \end{tabular}
        \\ 
      }
    \vspace{0.35in}
    \end{minipage}
  }
\caption{Grounding graph for the phrase ``put the pallet on the
  truck.''\label{fig:gg_pallet_on_truck}.  Both the entity and
  relation candidates have been simplified, as in general many more
  candidates are considered.  The best decoding result would result in
  $(\gamma_1,\gamma_2,\gamma_3,\gamma_4)\in\left\{(11,5,10,1),(12,6,9,8))\right\}$
  having high probability.}
  \label{figure:overview}
\end{figure*}

In $G^3$, each factor $\psi_m$ is trained as a binary classifier that
predicts which assignments to the grounding variables make sense for
each linguistic constituent.  These binary classes are introduced via
a correspondence variable $\phi_i\in\left\{0,1\right\}$.  When
$\phi_i=1$ for a linguistic constituent, then the probabilistic
predicate is true; when $\phi_i=0$, then it is false.  For example,
the probabilistic predicate for ``the pallet'' is true for object $3$,
but not for object $1$ in Figure~\ref{figure:environment:M}.  The
resulting random variables and factors for each linguistic constituent
in the parse tree therefore factor as follows:

\begin{align}
p(\Gamma | \Lambda, \Phi, M) = \frac{1}{Z} \prod_m \psi_m(\phi_m , \lambda_m, \gamma_{m_1} \dots \gamma_{m_k}).
\label{eq:factored_problem_statement}
\end{align}


The factorization in Equation~\ref{eq:factored_problem_statement} is
the core of the grounding graph; we can infer the grounding of
linguistic constituents independently, and the likelihood of the
entire text is the the joint likelihood of the independent factors. We
ensure that we can preserve needed correlations between groundings by
introducing two types of factors $\psi$. Entity factors are those that
predict the assignment of a constituent to a single specific grounding
($\psi(\phi_i, \lambda_i, \gamma_{m_1})$), typically a noun grounded
to some object or other discrete concept in the world.  In contrast,
relational factors predict the correspondence of a constituent across
two groundings (e.g., $\psi(\phi_i, \lambda_i, \gamma_{m_1},
\gamma_{m_2})$).  Relational factors can have arity greater than two
($\psi(\phi_i, \lambda_i, \gamma_{m_1}, \gamma_{m_2}, \gamma_{m_3})$)
and are mainly used for multi-argument verbs, such as
``put.''\footnote{All of the factors in this work were either entity
  factors or relational factors with arity 3 or less.}.  In both
cases, the individual factors $\psi_m$ quantify the correlation
between words in the linguistic constituent $\lambda_m$ and the
groundings $\gamma_{m_1} \dots \gamma_{m_k}$; the factors should have
large values where words correspond well to groundings and small
values otherwise.

The specific entity and relational factors $\psi_m$ are created
automatically for each sentence depending on the syntactic type of
linguistic constituents obtained from parsing that sentence, and the
parametric form of the $\psi_m$ will vary from application to
application. For example, for the application of mobile manipulation
(section \ref{sec:mobile_manipulation}), we learn the $\psi_m$ as
logistic regression classifiers, but for the application of route
following (Section~\ref{sec:navigation}), we learn the $\psi_m$ as
category distributions. We discuss in each section how we chose the
different correspondence models.

For example, Figure~\ref{fig:gg_pallet_on_truck} shows the graphical
model for the phrase ``Put the pallet on the truck.''  There are two
entity grounding variables $\gamma_2$ and $\gamma_4$ for the noun
phrases ``the pallet'' and ``the truck'' and values for these
variables range over objects in the environment.  Note that a
$\gamma_i$ could be referenced by more than one phrasal constituent;
this translates to being connected to more than one factor, as we see
that the value of $\gamma_4$ affects both the factor $\psi_4$ for
``the truck'' and the factor $\psi_3$ for ``on''.  However, the value
for $\gamma_4$ does not affect the grounding for $\psi_2$, ``the
pallet'' --- the model relies on the ``on'' factor to ensure that the
object grounded to ``the pallet'' is the same object that is on ``the
truck'', and that it is indeed on the truck.  Such independence
assumptions enable efficient inference and learning: our approach can
learn a model for the word ``on'' that can be used whenever this word
is encountered, without needing to manually introduce
a symbolic predicate for this concept.  We
describe how to formally derive the structural independencies from the
parse tree in Algorithm~\ref{fig:graph_algorithm}.

In order to enable efficient learning of generalizable predicates, we
use locally normalized factors for $\psi_m$.  Specifically:
\begin{align}
\psi_m(\phi_{m}, \lambda_m, \gamma_{m_1} \dots \gamma_{m_k})  \equiv p(\phi_m | \lambda_m, \gamma_{m_1} \dots \gamma_{m_k}). 
\end{align}

We can then write the factored form: 
\begin{align}
\label{eq:g3_factored}
p(\Gamma | \Lambda, \Phi, M) = \prod_m p(\phi_m | \lambda_m, \gamma_{m_1} \dots \gamma_{m_k}).
\end{align}

This factorization defines a probabilistic graphical model that
constitutes the grounding graph.  It is equivalent to
Equation~\ref{eq:factored_problem_statement}, except for a constant,
but assumes that the factors take a specific, locally normalized form.
Each factor $p(\phi_m | \lambda_m, \gamma_{m_1} \dots \gamma_{m_k})$
can be normalized appropriately --- only over the domain of the
$\phi_m$ variable for each phrasal constituent, substantially
improving our ability to learn the model.  Moreover, although each
factor roughly corresponds to a probabilistic predicate, these
predicates are defined by the linguistic structure.  The arity of the
predicate is defined by the number of children in the parse tree for
that node.  This representation is consistent with logical
representations for word meanings used in the literature on
semantics~\citep{zettlemoyer2005lms}.  The application of factors
$\psi$ results in set of scored groundings for each probabilistic
predicate, as in Figure~\ref{figure:overview:parsing}.  Based on these
scored candidates and the inferred linguistic structure, a grounding
graph is created (Figure~\ref{figure:grounding_graph_ex}) to represent
the constraints that exist between the different groundings.  During
inference, a configuration of $\Gamma$ is scored under the assumption
that $\phi_i=1$, which is equivalent to the notion of virtual evidence
nodes~\citep{bilmes2004virtual}~\citep{li2009use} in a conditional
random field.  Our approach learns these predicates during training;
because they are locally normalized they can be recombined through
Equation~\ref{eq:g3_factored} at test time in order to interpret
commands that have not been previously encountered.


Formally, a grounding graph consists of random variables for the
command $\Lambda = \left[\lambda_1 \dots \lambda_M\right]$, the
grounding variables $\Gamma = \left[\gamma_1 \dots \gamma_M\right]$, and the
correspondence variables $\Phi = \left[\phi_1 \dots \phi_M\right]$,
along with a set of factors $\Psi$.  Each factor $\psi \in \Psi$
consists of $\phi_m, \lambda_m, \gamma_{m_1} \dots \gamma_{m_k}$.
Algorithm~\ref{fig:graph_algorithm} specifies how random variables and
factors are automatically created from the parse structure of a
natural language command. 

\begin{figure}[t]
\centering
\subfigure[``Put the pallet on the truck.''\label{fig:ex_put}]{
\begin{tabular}{c}
\fbox{\parbox{0.45\linewidth}{
\tiny\tt 
(ROOT (S (VP (VB Put) \newline
\hspace*{0.31in}~~~~~~(NP (DT the) (NN pallet))\newline
\hspace*{0.31in}~~~~~~(PP (IN on)\newline
\hspace*{0.31in}~~~~~~~~~~(NP (DT the) (NN truck))))\newline
\hspace*{0.01in}~(. .)))
\newline
}}
\\
\fbox{\parbox{0.45\linewidth}{\centering\includegraphics{crf_put-crop.pdf}}}
\end{tabular}
}\subfigure[``Go to the pallet on the truck.''\label{fig:ex_go}]{
\begin{tabular}{c}
\fbox{\parbox{0.45\linewidth}{
\tiny \tt 
(ROOT (S (VP (VB Go)\newline
\hspace*{0.01in}~~~~~~~~~(PP (TO to)\newline
\hspace*{0.01in}~~~~~~~~~~~~~(NP (NP (DT the) (NN pallet))\newline
\hspace*{0.01in}~~~~~~~~~~~~~~~~~(PP (IN on)\newline
\hspace*{0.01in}~~~~~~~~~~~~~~~~~(NP (DT the) (NN truck))))))\newline
\hspace*{0.01in}~(. .)))
}}\\
\fbox{\parbox{0.45\linewidth}{\centering\includegraphics{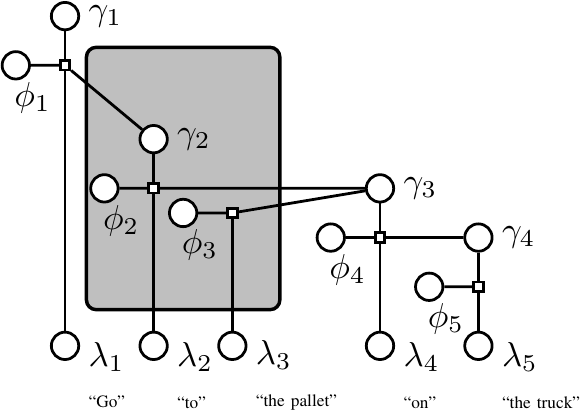}}}
\end{tabular}
}

\caption{Parse tree and induced model for two different commands.  The
  shaded region shows where the factorization differs.  \label{fig:ex}}

\end{figure}

Because the factors themselves are learned, this graphical structure
enables our approach to learn predicates for words such as ``on'',
``to,'' ``near'' and ``next to'' from lower-level features, unlike in
previous statistical approaches where these predicates must be
provided by the system designer.  These learned predicates can be
reused in commands not seen during training. Furthermore, this model
allows us to perform inference over similar sentences that have very
different meaning and parse structure. For example,
Figure~\ref{fig:ex} shows the grounding graphs for two different
natural language commands.  Figure~\ref{fig:ex_put} shows the parse
tree and graphical model generated for the command ``Put the pallet on
the truck.''  The random variable $\phi_2$ is associated with the
constituent ``the pallet'' and the grounding variable $\gamma_2$.  The
random variable $\phi_1$ is associated with the entire phrase, ``Put
the pallet on the truck'' and depends on both the grounding variables
$\gamma_1$ (the action that the robot takes) their arguments
$\gamma_2$ (the object being manipulated) and $\gamma_3$ (the target
location).  The $\lambda_i$ variables correspond to the text
associated with each constituent in the parse tree.
Figure~\ref{fig:ex_go} shows the parse tree and induced model for a
different command, ``Go to the pallet on the truck.'' Although the
words are almost the same between the two examples, the parse
structure is different, yielding a different graphical structure,
highlighted in gray, and a different grounding.

\renewcommand{\algorithmicrequire}{\textbf{Input:}}
\renewcommand{\algorithmicensure}{\textbf{Output:}}

\begin{algorithm}
\fbox{\parbox{1\linewidth}{
\begin{algorithmic}[1]
  \Require \\ 
  Parsed natural language command, $\Lambda = \lambda_1
  \dots \lambda_M$, with root $\lambda_{root} \in \Lambda$. 
  $\lvert N\rvert$ is the number entity factors.  
  $M=\lvert\Lambda\rvert$.
\State $\Phi \gets \phi_1 \dots \phi_M$ 
\State $\Gamma \gets \gamma_1 \dots \gamma_N$ 
\State $\Psi \gets []$
\For {$\lambda_i \in \Lambda$}
  \State {\bf If $\lambda_{i}$ is entity:}
  \State Add $(\phi_i, \lambda_i, \gamma_{i})$ to $\Psi$
  \newline

  \State {\bf If $\lambda_{i}$ is relation:}
  \State $\Gamma_{factor}  \gets []$
  \For {each $\lambda_{j} \in args(\lambda_i)$}
  \State Add $\gamma_{j}$ to $\Gamma_{factor}$ 
  \EndFor
  \State Add $\gamma_{j}$ to $\Gamma_{factor}$ for each implicit argument for $\lambda_i$

  \State Add $(\phi_i, \lambda_i, \Gamma_{factor})$ to $\Psi$
  \newline 

\EndFor
  \Ensure $\Lambda, \Phi, \Gamma, \Psi$
\end{algorithmic}
}}
\caption{Generating a grounding graph from natural language command.
  $args$ is a function that determines the arguments to a relational
  linguistic constituent (usually this function just gets the children
  of $\lambda_i$).  For arguments to the relation that are implicit
  (but necessary for understanding), we add implicit groundings
  $\gamma_j$.
  \label{fig:graph_algorithm}}
\end{algorithm}

\section{Mobile Manipulation}
\label{sec:mobile_manipulation}
\begin{figure}
\centering \subfigure[Robotic forklift, operating in an outdoor supply
  depot.\label{fig:forklift}]{\includegraphics[width=0.49\linewidth]{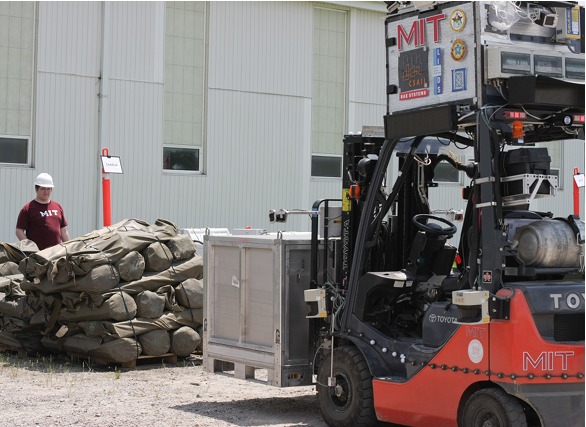}}\hfil%
\subfigure[PR2, a mobile humanoid operating in indoor, household
  environments.\label{fig:pr2}]{\includegraphics[width=0.49\linewidth]{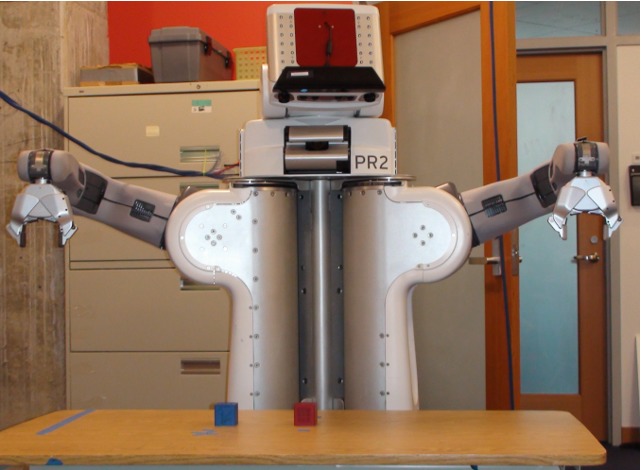}}
\caption{We demonstrated our command-understanding framework on two
  mobile-manipulation robots.\label{fig:mobileManipulationRobots}}
\end{figure}

Robots capable of manipulating objects can prove tremendously useful
to human partners.  Example platforms include a robotic forklift
(Figure~\ref{fig:forklift}), designed for outdoor, large-scale
manipulation~\citep{teller10}, and the PR2 (Figure~\ref{fig:pr2}), designed for indoor,
household tasks).  For example, the autonomous forklift might be tasked
with moving pallets in a supply depot in preparation for offloading or
distribution.  The PR2 might carry out tasks in the home, such as unpacking
and putting away a shipment of household products.  In both cases, a user might wish to
specify where the objects should go, as in ``Put the pallet in
receiving,'' or ``Hand me the stapler.''  This section describes how
we use the \G3 framework to understand natural language commands for
robots like these.  We present a quantitative evaluation in simulation
for the forklift domain, and demonstrate the end-to-end system on both
the forklift and the PR2.

\subsection{Modeling Word Meanings} 

\label{sec:mm_word_meanings}

The specific form for each factor $p(\phi_m | \lambda_m, \gamma_{m_1}
\dots \gamma_{m_k})$ varies in our different application domains.
For mobile manipulation, we assume that each factor in
Equation~\ref{eq:g3_factored} takes a log-linear form with feature
functions $s_j$ and feature weights $\theta_j$:
\begin{align}
p(\phi_m | &\lambda_m,  \gamma_{m_1} \dots \gamma_{m_k})= \frac{1}{Z} \exp\left(\sum_j \theta_j s_j(\phi_m, \lambda_m, \gamma_{m_1} \dots \gamma_{m_k})\right).
\label{eq:loglinear}
\end{align}

Word meanings are represented as weights associated with these feature
functions.  The specific features capture the mapping between words in
language and aspects of the external world and are described in more
detail in Section~\ref{sec:mm_word_meanings}.  We optimize feature
weights $\theta_j$ to maximize the likelihood of the training data.
This function is convex and can be optimized with gradient-based
methods, such as L-BFGS~\citep{andrew07, mallet}.

We train the system using data that consists of natural language
commands together with positive and negative examples of groundings
for each constituent in the command.  The robot was given information
about the environment, including all objects in the environment with
their labels.  For example, the forklift model was given information
about all pallets in the environment, their tag as ``pallet'' as well
as their location and geometry (as well as other objects such as the
truck). We manually annotated the alignment between nouns in the
corpus that corresponded to an object, place, path, or event sequence
in the external world. By contrast, verb phrase grounding was
automatically aligned with an agent path or event from the log
associated with the original video.  This automatic annotation
substantially reduced the annotation effort for verb phrases, but was
an approximation
that admittedly led to inaccurate alignments for compound commands
such as ``Pick up the right skid of tires and place it parallel and a
bit closer to the trailer,'' where each verb phrase refers to a
different part of the state sequence.  This lack of perfect alignment
introduced noise into the training process.  In the future we plan to
explore automatically segmenting the trajectories and associating each
part of the trajectory with the appropriate part of the verb phrase.
However this problem is challenging because trajectories are
continuous in time.  

The annotations above provided positive examples of grounded language.
In order to train the model, we also need negative examples.  We
generated negative examples by associating a random grounding with
each linguistic constituent.  Although this heuristic works well for
verb phrases, ambiguous noun phrases such as ``the pallet'' or ``the
one on the right'' are often associated with a different, but still
correct, object (in the context of that phrase alone).  For this
reason, we manually corrected the negative noun phrase examples,
reannotating some of them as positive.

We define binary features $s_k$ for each factor.  These features
enable the system to determine which values for $\Gamma$ correctly
ground the corresponding linguistic constituent.  Geometric ``base''
features enable the system to represent relations involving objects,
places, paths, and events.  For a relation such as ``on,'' a natural
geometric feature is whether the first argument is supported by the
second argument, taking into account their geometric relationship.
However, the base feature $supports(\gamma_i^f, \gamma_i^l)$ alone is
not enough to enable the model to learn that ``on'' corresponds to
$supports(\gamma_i^f, \gamma_i^l)$, because this feature is
independent of the language.  Instead we need a feature like
$supports(\gamma_i^f, \gamma_i^l) \wedge \mbox{``on''} \in
\lambda_i^r$.  Thus, the system creates the Cartesian product of base
features and the words in the corresponding linguistic constituents to
compute features $s_k$. Base features that are continuous-valued
functions such as $DistanceTo(\gamma_i)$ are discretized to create a
set of binary base features.

We manually implemented a set of base features that involve geometric
relations between the groundings $\gamma_i$.  Since groundings for
prepositional and verb phrases correspond to the location and
trajectory of the robot and any objects it manipulates over time, the
feature functions for these linguistic constituents require 
as input the geometry of the robot, the object it is manipulating, as well as their trajectories through space.
Examples include:
\begin{itemize}
\item The displacement of a path toward or away from a landmark object.
\item The average distance of a path from a landmark object.
\end{itemize}

We also used the complete set of features described in
\citet{tellex10a}, which capture concepts such as contact, and the
relative geometric positions between two objects.  We use 49 base
features for leaf noun phrases (e.g., ``the truck'') and prepositional
phrases (e.g., ``on the truck''), 56 base features for compound noun
phrases (e.g., ``the pallet on the truck''), and 112 base features for
verb phrases (e.g., ``Pick up the pallet.'').  Automatically
discretizing features and taking the Cartesian product with words from
the training set produces 147,274 binary features.  These features are
defined in the Appendix~\ref{sec:appendix}.

\subsection{Inference}

\begin{figure*}[t]
\centering \subfigure[Object
  groundings.]{\begin{overpic}[scale=0.25,tics=10,width=0.32\linewidth]{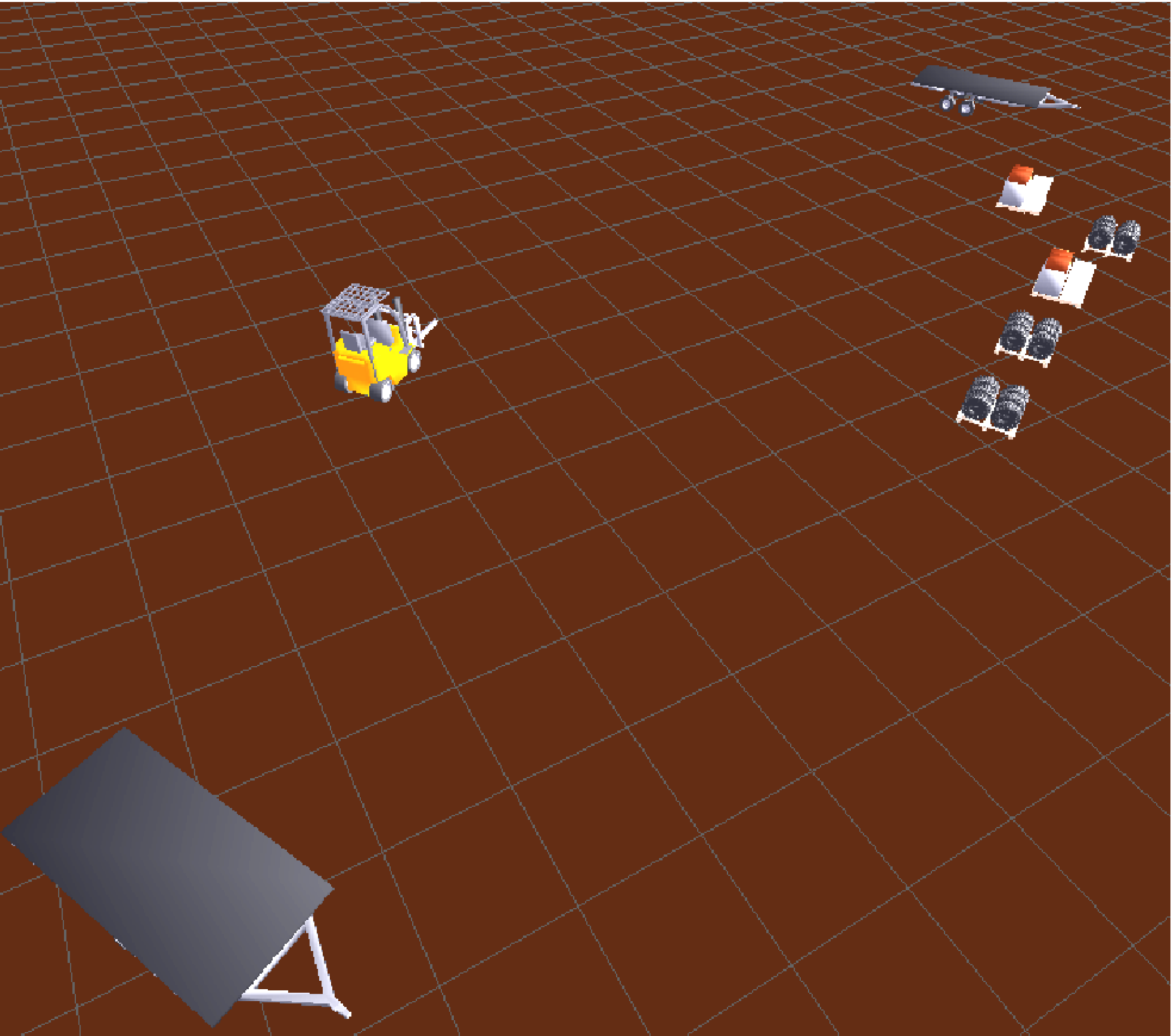}
      \put(15, 15){\color{white}{\circle{15}}}
      \put(24, 9){\color{white}{\tiny Grounding for $\gamma_4$}}
      \put(17, 18.5){\color{white}{\circle{6}}}
      \put(20, 23){\color{white}{\tiny Grounding for $\gamma_3$}}
      \put(85, 53.5){\color{white}{\circle{9}}}
      \put(50, 46){\color{white}{\tiny Grounding for $\gamma_2$}}
\end{overpic}}
\hfill
\subfigure[Pick up the pallet.]{\begin{overpic}[scale=0.25,tics=10,width=0.32\linewidth]{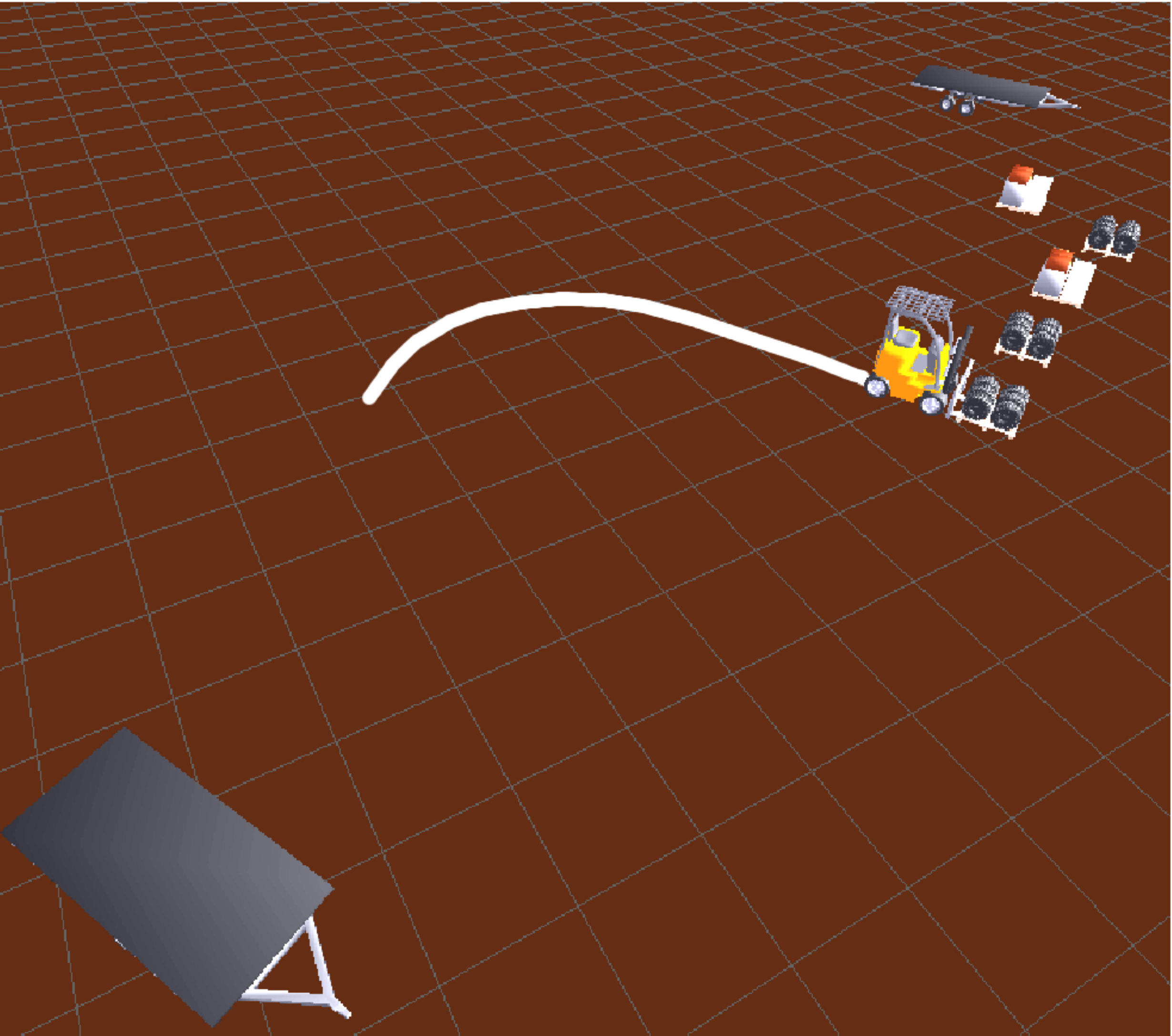}
%
%
%
%
\end{overpic}}
\hfill
\subfigure[Put it on the truck.]{\begin{overpic}[scale=0.25,tics=10,width=0.32\linewidth]{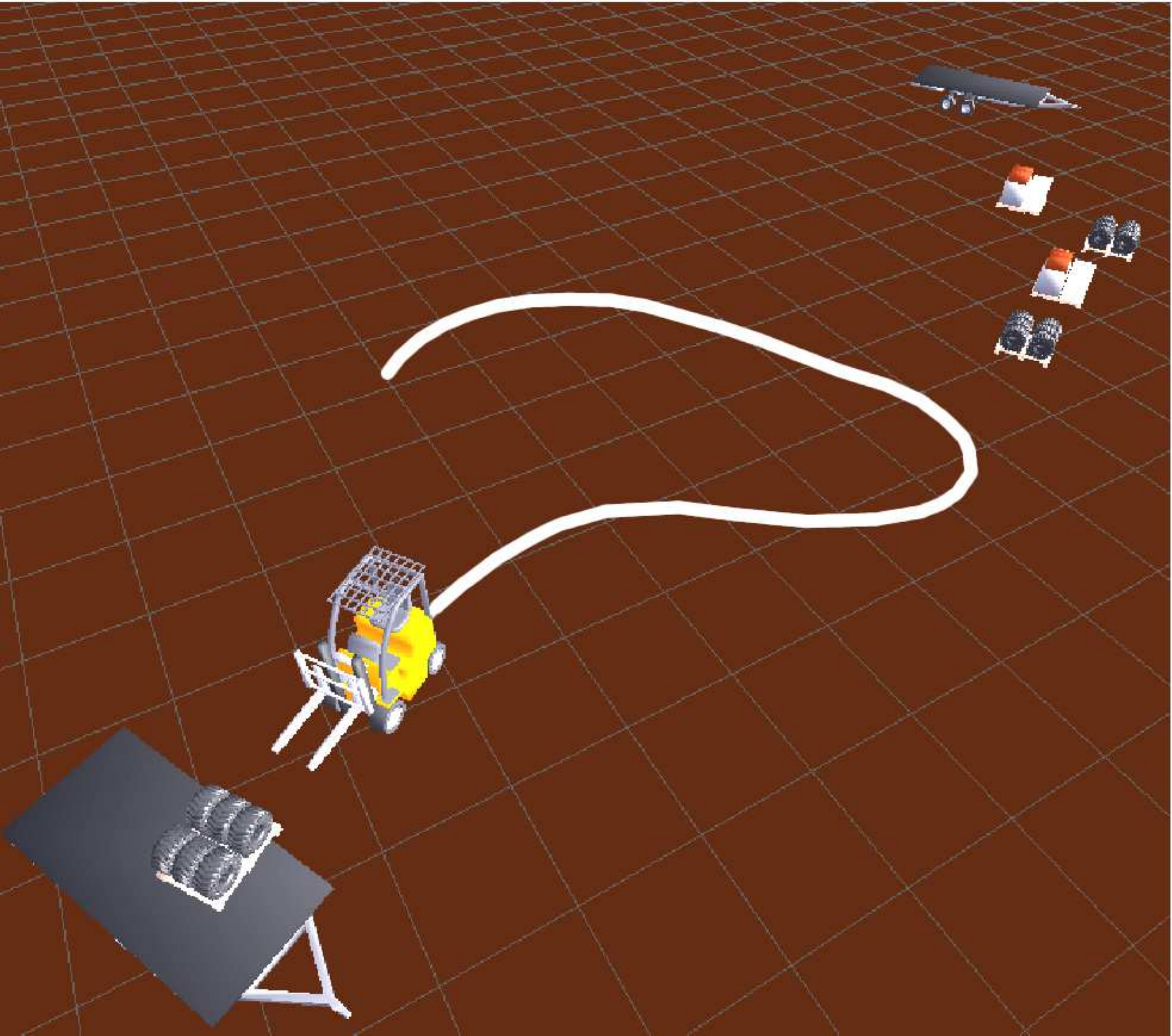}
\put(50, 35){\color{white}{\tiny Grounding for $\gamma_1$}}
\end{overpic}}
\caption{A sequence of the actions that the forklift takes in
  response to the command ``Put the tire pallet on the truck.''  In
  (a) the search grounds objects and places in the world based on
  their initial positions.  In (b) the forklift executes the first
  action and picks up the pallet.  In (c) the forklift puts the
  pallet on the truck.\label{fig:command}}
\end{figure*}
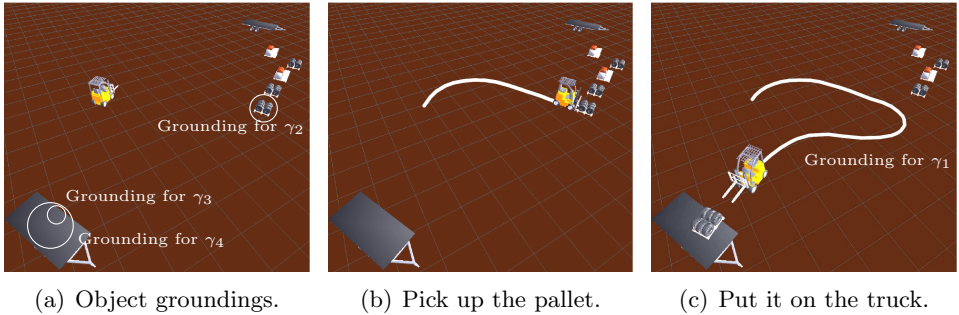

Given a command, we want to find the set of most probable groundings
for that command.  During inference, we search for groundings $\Gamma$
that maximize the likelihood of the groundings for each linguistic
constituent as in Equation~\ref{eq:factored_problem_statement}. To
limit the search space, the system uses a topological map of the
environment that defines a limited set of salient objects and
locations. In our experiments, the map size ranged between
approximately three and twenty objects and locations.  Using the
topological map, we define a discrete state/action space for the
robot, and search for sequences of actions corresponding to the
grounding for verb phrases.  Even though we discretize the search
space, our optimization considers all permutations of object
assignments as well as every feasible sequence of actions the agent
might perform.  As a result, the search space becomes large as the
number of objects, paths, and events in the world increases.  In order
to make the inference tractable, we use beam search with a fixed beam
width.  For noun phrases and place-prepositional phrase groundings,
the beam width was $10$; for verb phrase and path-prepositional
phrases, the beam width over candidate state sequences was $5$.  We
determined the beam width empirically from data, finding the narrowest
beam width that led to good performance.  Figure~\ref{fig:command}
shows the results of inference for an example from the forklift
domain; it consists of groundings for each linguistic constituent in
the command along with an action trajectory for the robot.

\subsection{Evaluation}

We collected a corpus of mobile manipulation commands paired with
robot actions and environment state sequences.  We used this corpus to
train the \G3 framework and also to evaluate end-to-end performance of
the system at following realistic commands from untrained users.  To
collect commands, we posted videos of action sequences to Amazon's
Mechanical Turk (AMT) and collected language associated with each
video.  The videos showed a simulated robotic forklift engaging in an
action, such as picking up a pallet or moving through the environment.
Paired with each video, we collected a complete log of the state of
the environment and the robot's actions.  Subjects were asked to type
a natural language command that would cause an expert human forklift
operator to carry out the action shown in the video.  We collected
commands from 45 subjects for twenty-two different videos.  Each
subject interpreted each video only once, but we collected multiple
commands (an average of 13) for each video, for a total of 285
commands.  The corpus contained 6149 words, with 508 unique words.
Figure~\ref{fig:mm_commands} shows example commands from our dataset,
which the system can successfully follow.

\begin{figure}
\fbox{\parbox{1\linewidth}{
Go to the first crate on the left and pick it up.\\

Pick up the pallet of boxes in the middle and place them on the
trailer to the left.\\

Go forward and drop the pallets to the right of the first set of
tires.\\

Pick up the tire pallet off the truck and set it down.\\
}}
\caption{Example commands from our corpus, which the system can successfully follow.\label{fig:mm_commands}}
\end{figure}


Subjects were not primed with any example words or phrases from the
domain describing the actions or objects in the video, leading to a
wide variety of natural language commands including nonsensical ones
such as ``Load the forklift onto the trailer,'' and misspelled ones
such as ``tailor'' (trailer).  Figure~\ref{fig:exampleForklift} shows
commands collected using this methodology for one video in our
dataset.

\subsubsection{Model Evaluation}


We annotated each constituent in the corpus with the corresponding
grounding.  Using the annotated data, we trained the model and
evaluated its performance on a held-out test set.  We did a single
random training/test split with $70\%$ in training and $30\%$ in test.
We used one set of videos (with associated commands) in the training
set, and a different set in the test set.  There was no development
set used to validate hyperparameters.  We assessed the model's
performance at predicting the correspondence variable given access to
words in the language and ground truth values for the grounding
variables.  The test set pairs a disjoint set of scenarios from the
training set with language given by subjects from AMT.  This process
evaluates Equation~\ref{eq:loglinear} directly;
Section~\ref{sec:mm-e2e-eval} conducts an end-to-end evaluation.

\begin{table}[h]
\centering
{\small
\begin{tabular}{rcccc}
\toprule
Constituent type  & Precision & Recall & F-score & Accuracy \\
\midrule
Noun Phrase & 0.93 & 0.94 & 0.94 & 0.91 \\
Prepositional Phrase (Place) & 0.70 & 0.70 & 0.70 & 0.70 \\
Prepositional Phrase (Path) & 0.86 & 0.75 & 0.80 & 0.81 \\
Verb Phrase & 0.84 & 0.73 & 0.78 & 0.80 \\
\midrule
Overall & 0.90 & 0.88 & 0.89 & 0.86 \\
\bottomrule
\end{tabular}
}
\caption{Performance of the learned model at predicting the
  correspondence variable $\phi$.\label{crf:performance}}
\end{table}

Table~\ref{crf:performance} reports overall performance on this test
set and performance broken down by constituent type.  The performance
of the model on this corpus indicates that it robustly learns to
predict when constituents match groundings from the corpus.  We
evaluated how much training was required to achieve good performance
on the test dataset and found that the test error asymptotes at around
1,000 (of 3,000) annotated constituents.

For noun phrases, correctly-classified high-scoring examples in the
dataset include ``the tire pallet,'' ``tires,'' ``pallet,'' ``pallette
[\emph{sic}],'' ``the truck,'' and ``the trailer.''  Low-scoring
examples included noun phrases with incorrectly annotated groundings
that the system actually got right.  A second class of low-scoring
examples arose due to words that appeared rarely in the corpus.


For place prepositional phrases, the system often correctly classified
examples involving the relation ``on,'' such as ``on the trailer.''
However, the model often misclassified place prepositional phrases
that involve frame-of-reference.  For example, ``just to the right of
the furthest skid of tires'' requires the model to have features for
``furthest,'' which requires a comparison to other possible objects
that match the phrase ``the skid of tires.''  Understanding ``to the
right'' requires reasoning about the location and orientation of the
agent with respect to the landmark object.  Similarly, the phrase
``between the pallets on the ground and the other trailer'' requires
reasoning about multiple objects and a place prepositional phrase that
has two arguments.  The model is not capable of interpreting phrases
like ``the first'' or ``second'' crate on the left, because it cannot
handle subsets or groups of objects.  (Although our more recent work has addressed this limitation~\citep{paul16}.)

For verb phrases, the model generally performed well on ``pick up,''
``move,'' and ``take'' commands.  The model correctly predicted plans
for commands such as ``Lift pallet box,'' ``Pick up the pallets of
tires,'' and ``Take the pallet of tires on the left side of the
trailer.''  It predicted incorrect plans for commands like, ``move
back to your original spot,'' or ``pull parallel to the skid next to
it.'' The word ``parallel'' appeared in the corpus only twice, which
was apparently insufficient to learn a good model.  ``Move'' had few
good negative examples, since we did not have in the training set
contrasting examples of paths in which the forklift did not move.

\subsubsection{End-to-end Evaluation}
\label{sec:mm-e2e-eval}
To evaluate end-to-end performance, the system inferred plans given
only commands from the test set, a starting location for the robot and
a map of the environment with the location and type of objects located
in the map.  We segmented commands containing multiple top-level verb
phrases into separate clauses.  Next, the system used the generated
grounding graph to infer a plan and a set of groundings for each
clause.  We simulated plan execution on a realistic, high-fidelity
robot simulator from which we created a video of the robot's
actions. We uploaded these videos to Amazon Mechanical Turk
(AMT)\footnote{www.mturk.com} where subjects viewed each video paired
with a command and reported their agreement with the statement, ``The
forklift in the video is executing the above spoken command'' on a
five-point Likert scale.  We report command-video pairs as correct if
the subjects agreed or strongly agreed with the statement, and
incorrect if they were neutral, disagreed or strongly disagreed.  We
collected five annotator judgments for each command-video pair.

To validate our evaluation strategy, ensure that the collected data is
interpretable and that a Likert evaluation metric can correctly
characterize good and bad grounding performance, we gave known correct
and incorrect command-video pairs to subjects on AMT.  In the case of
known correct grounding, subjects saw videos with known-correct
commands that were generated by other subjects. In the case of known
incorrect grounding, the subject saw the command paired with a random
video that was not used to generate the original command.
Table~\ref{tab:top_30} depicts the percentage of command-video pairs
deemed consistent for these two conditions.  As expected, there is a
large difference in Likert score between commands paired with the
original and randomly selected videos, validating our approach to
evaluation.  Additionally, these results show that commands in the
corpus are generally understandable by a different annotator but that
some people did give commands that other people found difficult to
follow.


We then evaluated our system by considering three different
configurations. Serving as a baseline, the first experimental
evaluation consisted of ground truth parse trees and a cost function
which selected actions at random. The second configuration involved
ground truth parse trees, and a learned cost function that selects the
best action to follow the commands. The third consisted of
automatically extracted parse trees and a learned cost function for
selecting the best action.


Table~\ref{tab:top_30} reports the performance of each configuration
along with their 95\% confidence intervals.  We evaluated on all
commands in the test set.  Additionally, as a second trial, we report
results on a subset of commands that were ranked according to the
model's own confidence score ({\it high conf}).  The relatively high
performance of the random cost function configuration relative to
commands paired with random videos demonstrates the inherent knowledge
captured by the discretized state/action space.  However, in all
conditions, the system performed statistically significantly better
than a random cost function. The system qualitatively produced
compelling end-to-end performance.  Even when it made a mistake, it
often correctly followed parts of the command.  For example, it
sometimes picked up the left-hand tire pallet rather than the
right-hand pallet.  Other types of errors arose due to ambiguous or
unusual language, such as ``remove the goods'' or ``the lonely
pallet.''

The system performed noticeably better on the high confidence commands
than on the entire test set.  This result indicates the validity of
our probability measure, suggesting that the system had some knowledge
of when it is correct and incorrect.  We have demonstrated that the
system can use this information to decide when to ask for confirmation
before acting~\citep{tellex13, deits13}.

\begin{table}[t]
\centering
\begin{tabular}{rl}
\toprule
\multicolumn{1}{c}{Scenario} & \% Correct\\
\midrule
Commands paired with original video & $91\% \pm 1\%$ \\
Commands paired with random video & $11\% \pm 2\%$ \\
Annotated parses (high conf.), learned cost & $63\% \pm 8\%$\\
Automatic parses (high conf.), learned cost & $54\% \pm 8\%$ \\
Annotated parses (all), learned cost & $47\% \pm 4\%$ \\
Annotated parses (all), random cost  & $28\% \pm 5\%$ \\
\bottomrule
\end{tabular}
\caption[End-to-end performance in different scenarios.]{The fraction
  of commands considered correct by our annotators for different
  configurations. High conf. evaluated in the top 30 most probable
  commands according to the model.  We report 95\% confidence
  intervals.\label{tab:top_30}}
\end{table}

\subsubsection{Real-world Demonstration}

\begin{figure}[t]
\centering
\subfigure[Photographs of the forklift executing the command,
  ``Put the tire pallet on the truck.'' \label{fig:forklift_video}]{%
\includegraphics[width=0.32\linewidth]{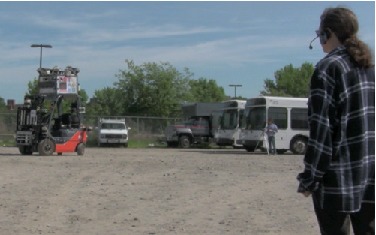}
\includegraphics[width=0.32\linewidth]{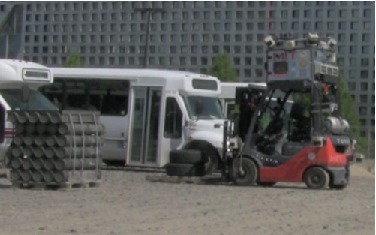}
\includegraphics[width=0.32\linewidth]{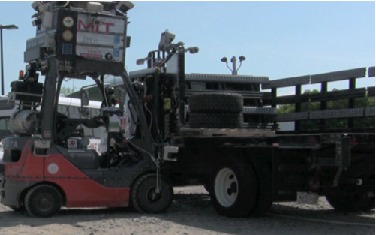}
}
\subfigure[Photographs of the PR2 executing the command, ``Pick up the red block.''\label{fig:pr2_video}]{%
\includegraphics[width=0.32\linewidth]{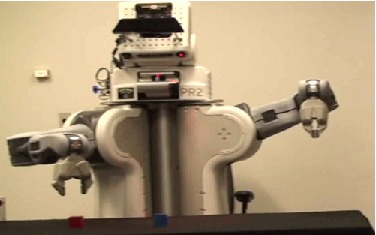}
\includegraphics[width=0.32\linewidth]{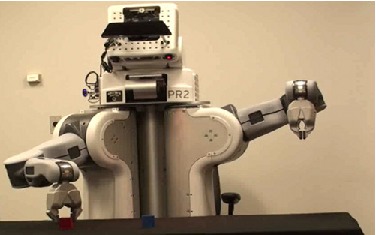}
\includegraphics[width=0.32\linewidth]{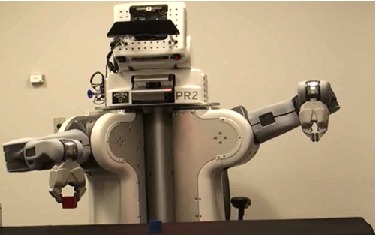}
}
\caption{Imagery from the PR2 and forklift following natural language
  commands using the \G3 framework.\label{fig:mobile_manipulation_interactive}}
\end{figure}

Finally, we demonstrated the end-to-end system on two platforms: a
robotic forklift and the PR2 mobile manipulator.  The robotic
forklift, described in detail by \citet{teller10}, is an autonomous
robotic vehicle capable of driving through real-world warehouse
environments.  It can localize itself, avoid obstacles, track people,
recognize objects, and move pallets.  Using the models described in
the previous section, we demonstrated that it can follow commands such
as ``Put the tire pallet on the truck'' and ``Pick up the tire
pallet.''  Figure~\ref{fig:forklift_video} shows scenes from a video
of the forklift executing the command ``Put the tire pallet on the
truck.''  Audio was captured using a headset microphone and converted
to text using the SLS recognizer~\citep{glass2003probabilistic}.
Next, the system parsed the text with the Stanford
Parser~\citep{de_marneffe_generating_2006} and extracted the graphical
model structure using Algorithm~\ref{fig:graph_algorithm}. The robot
had previously been given a tour of the environment that provided a
model of the visual appearance and label of salient objects, such as
the tire pallet and the truck, as described by~\citet{walter12}.
Finally, it carried out the inference in Equation~\ref{eq:g3_argmax}
to infer an action sequence and executed the action. See
\url{http://youtu.be/JyDRXOhr3b0} and
\url{http://youtu.be/OzWTyH4nGIc} for videos.

Using the PR2, we demonstrated command-following in a simple
blocks-world domain.  We used the Kinect sensor to detect the plane of
the table and segment the locations of blocks.  We collected a small
training corpus of natural language commands to train the model.  The
robot recognized speech using Google's speech recognizer, extracted
the graphical model from the text, and inferred groundings, including
an action for the robot.  We used {\tt rosbridge}~\citep{crick11} to
connect between the different systems.  Scenes from the robot
executing a command appear in Figure~\ref{fig:pr2_video}; see
\url{http://youtu.be/Nf2NHlTqvak} for video. 

Our real-world demonstrations tended to use simpler commands than the
corpus because of the difficulty and expense creating real-world test
environments.

\section{Route Directions}
\label{sec:navigation}
\begin{figure}
\centering
\subfigure[Robotic wheelchair.\label{fig:wheelchair}]{\includegraphics[height=1.5in]{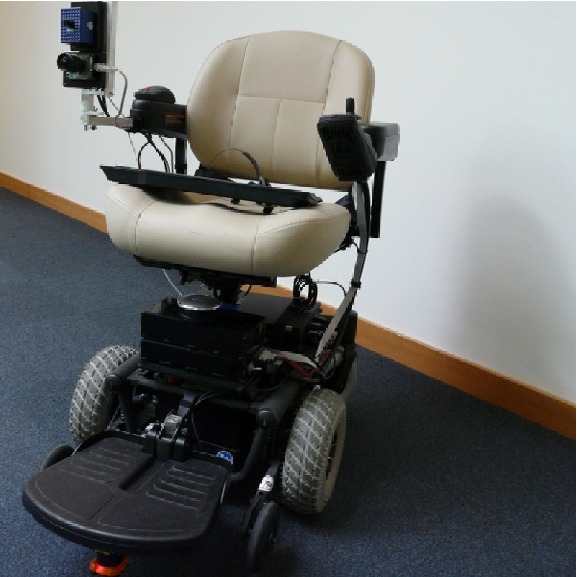}}
\subfigure[Robotic quad-rotor helicopter.\label{fig:mav}]{\includegraphics[height=1.5in]{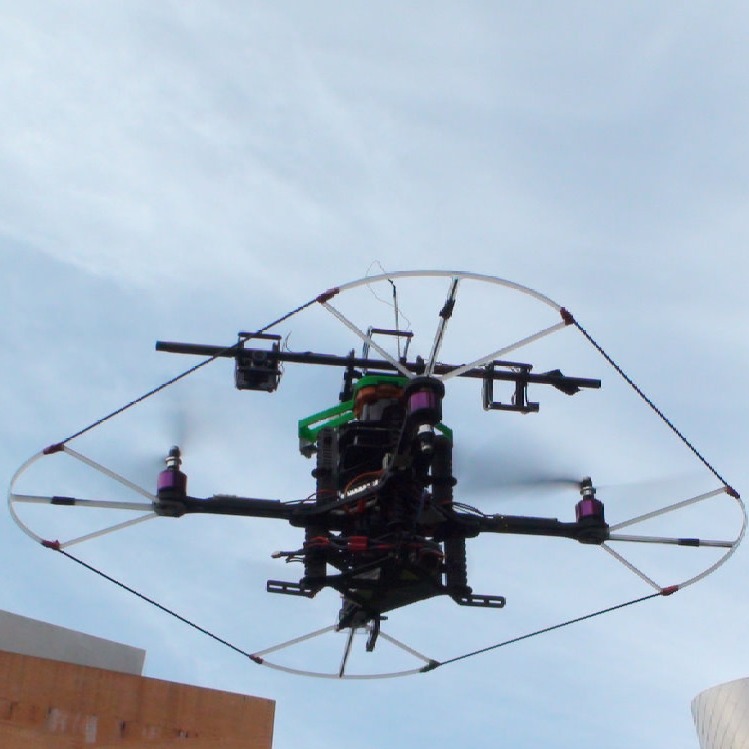}}
\caption{Two robots used in our direction following work.\label{fig:directionRobots}}
\end{figure}

In the previous section, we tested our system at following
mobile-manipulation commands on single verb phrases.  However, in many
task domains, it would be useful to understand longer movement
commands.  For example, Figure~\ref{fig:wheelchair} shows a robotic
wheelchair; understanding natural language movement commands would
enable a person to use language to control the chair, even if they
were unable to control it with a conventional interface.
Figure~\ref{fig:mav} shows a robotic micro-air vehicle (MAV), which
can engage in inspection tasks as well as search-and-rescue
operations.  Specifying a three-dimensional trajectory for such a
vehicle using conventional interfaces is challenging for untrained
users.  If instead, a user could speak a natural language command
describing where s/he wanted the vehicle to go, it would enable
higher-level, intuitive interaction.
Figures~\ref{fig:exampleWheelchair} and \ref{fig:exampleMav} show
examples of these types of longer instructions.  


As can be seen in these examples, typical route directions can result
in a long sequence of clauses that the robot must follow, making the
inference problem more challenging.  In order to infer the optimal set
groundings (e.g., paths, landmarks) for the sequential structure of
route directions, inference must consider vastly more paths through
the environment than in Section~\ref{sec:mobile_manipulation} because
commands are substantially longer.  To enable efficient and robust
inference, we make an approximation to the full hierarchical model by
creating a fixed, flat linguistic structure.  This structure not only
enables the search space over paths to become more tractable, we also
found that in practice the flat model is more effective at capturing
the meaning of route directions because they better tolerate
discrepancies in object groundings.  For example, if a robot is told
``Go to the lounge with couches'' and it maps the noun phrase ``the
lounge with couches'' to the couches instead of the lounge, it will
probably perform roughly the right action when it tries to go there.
Note that this kind of approximation generally does not hold for
mobile manipulation: if the person says, ``Pick up the tire pallet
near the box pallet'' and the robot picks up the box pallet instead of
the tire pallet, then it has not accurately followed the command.
This section describes how we perform efficient inference in a
flattened model to enable a robot to understand longer sequences of
natural language movement commands.
\begin{figure}[t]
\centering
%
\subfigure[Hierarchical grounding graph.\label{fig:parse_hierarchical}]{
\centering
\fbox{\parbox{0.95\linewidth}{
\centering
\includegraphics[height=1.6in]{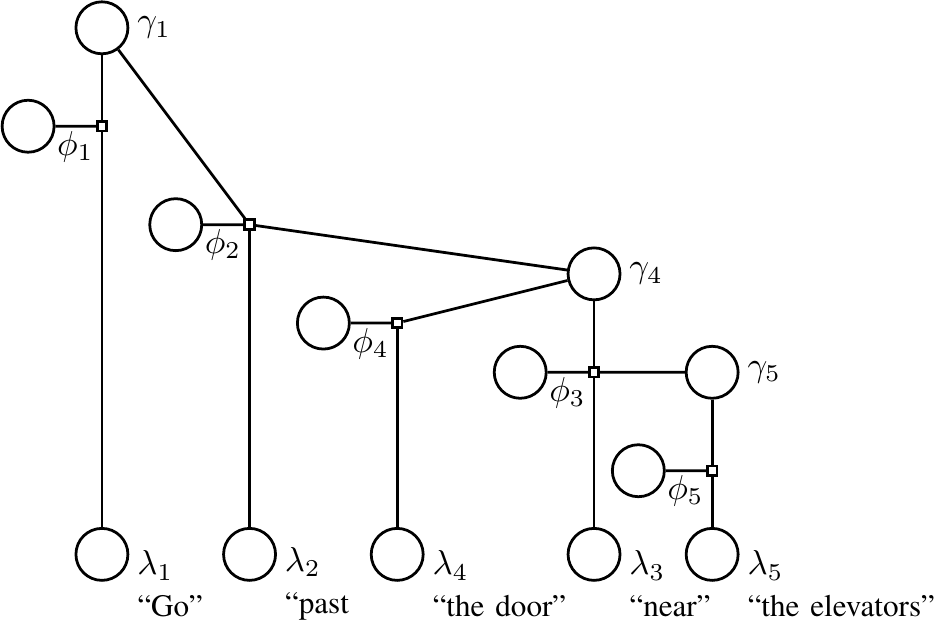}
}}}
%
%
%
\subfigure[Flattened grounding graph.\label{fig:parse_flattened}]{
\centering
\fbox{\parbox{0.95\linewidth}{
\centering
\includegraphics[height=1.6in]{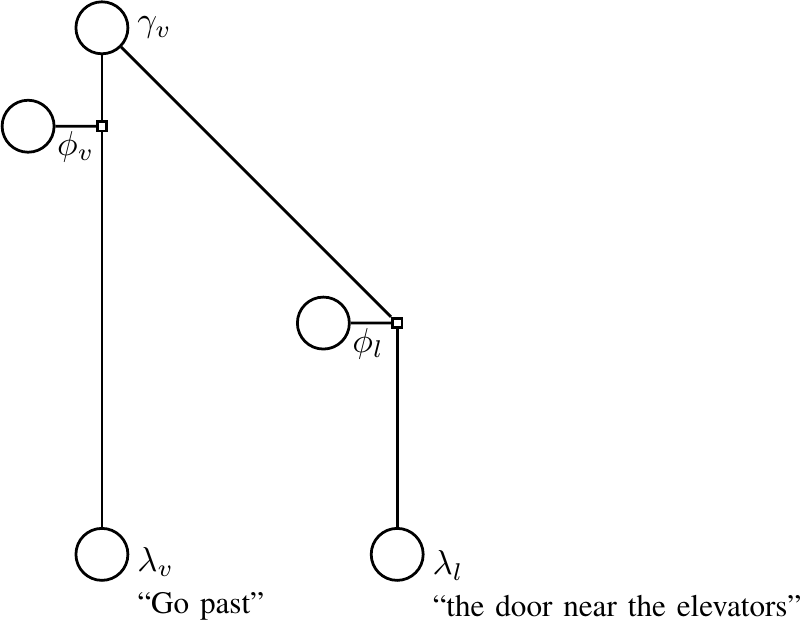}
}}}
%
\caption{Hierarchical and flattened grounding graphs for the sentence,
  ``Go past the door near the elevators.''
\label{fig:parses}}
\end{figure}

\subsection{Modeling Word Meanings}
\label{sec:rd_word_meanings}
To generate a grounding graph with a simplified structure, we use a
flattened parse tree extracted using a CRF
chunker~\citep{crfpp}.  Figure~\ref{fig:parses} shows a comparison of
the flattened and full hierarchical models for the same sentence.  The
flattened structure enables much more efficient inference using a
variant of the Viterbi algorithm for multiple sentences, so that the
system can quickly infer the actions implied by a paragraph-length set of
route instructions.

We assume that the command consists of a sequence of $S$ phrases,
where each phrase breaks down into linguistic constituents
$(\lambda_v,\lambda_l)$ and corresponding groundings
$(\gamma_v,\gamma_l)$.  This structure leads to efficient inference
algorithms for finding the optimal trajectory, described in the
following section.  Adapting the framework described in
Section~\ref{sec:g3} to this more simplified structure, there is a
correspondence variable for the landmark $\phi_v$ and one for the
landmark $\phi_l$, such that: 


\begin{align}
p(\Gamma | \Lambda, \Phi, M) &=
 \prod_s p(\phi_v | \lambda_v, \gamma_t) \times p(\phi_{l} | \lambda_{l}, \gamma_t).
\label{eq:factor_rd}
\end{align}
This factored form is equivalent to Equation~\ref{eq:g3_factored} but
is explicit about how factors are constructed from the command.  By
using a fixed factorization, we sacrifice linguistic flexibility in
order to simplify model learning and obtain efficient inference
algorithms that exploit the repeating sequential structure of the
factorization.  The grounding variable $\gamma_t$ corresponds to the
trajectory associated with this path segment, represented as $T$
points $s_1 \dots s_T$.  Note that the landmark factor $\phi_l$
connects directly to $\gamma_t$ without an intermediary variable
$\gamma_l$.

Because route directions generally use a restricted set of movement
actions but a more open-ended set of landmark objects, we use
different models for each factor in the grounding graph.  We connect
the landmark factor directly to the trajectory and estimate the
probability that an object corresponding to the landmark phrase can be
seen from the endpoint of the trajectory using object-object
co-occurrence statistics mined from a large dataset (described in
Section~\ref{sec:flickr}).  In addition, we model how well the
trajectory corresponds to a verb phrase in the command (described in
Section~\ref{sec:verb}).  Modeling more of the linguistic structure,
such as spatial relations, requires that the landmark be accurately
grounded to a specific object in the robot's model of the environment;
our previous work showed that explicitly modeling spatial relations
did not contribute significantly to overall
performance~\citep{kollar10}.

\subsubsection{Verbs}
\label{sec:verb}

To compute $p(\phi_v | \lambda_v, \gamma_t)$, that is, to ground verb
phrases, we use a three-state model that classifies a verb phrase
$\lambda_v$ as either ``left,'' ``right,'' (if $\lambda_v$ contains
the word ``left'' or ``right'') or ``straight'' (otherwise).  We
define the verb features precisely in the
Appendix~\ref{sec:appendix}.  This approach is able to interpret
verbs such as ``Go left,'' ``go right,'' ``move forward'' and the
like.

\subsubsection{Noun Phrases}
\label{sec:flickr}
Modeling noun phrases is challenging because people refer to a wide
variety of objects in natural language directions and use diverse
expressions to describe them.  In our corpus, people used more than
150 types of objects as landmarks, ranging from ``the door near the
elevators'' to ``a beautiful view of the domes.''  To compute
$p(\phi_l | \lambda_l, \gamma_l)$, that is, to ground landmark
phrases, the system takes a semantic map seeded with the locations of
known objects and uses object-object context to predict the locations
of the unknown landmark terms, following \citet{kollar09}.
Object-object context allows the system to predict that a computer is
nearby if it can directly detect a monitor and a keyboard, even if it
cannot predict the computer's exact location and geometry.  To predict
where a novel landmark may occur, we downloaded over a million images,
along with their associated labels from the photo-sharing website
Flickr.

We computed the probability that a particular word $w_j \in
\lambda_l$ applies to the trajectory $\gamma_t$, given the detected
object labels associated with the trajectory $O(\gamma_t)$:
\begin{align}
p(\phi_l | \lambda_l, \gamma_t) = p(\phi_l| w_1 \dots w_J, O(\gamma_t)).
\label{eqn:flickr}
\end{align}
Next, we rewrite the equation using Bayes' rule, assuming the words
$w_j \in \lambda_l$ are independent:
\begin{align}
p(\phi_l| w_1 \dots w_J, O(\gamma_t)) &= \frac{\prod_j p(w_j | \phi_l, O(\gamma_t)) p(\phi_l | O(\gamma_t))} {p(w_1 \dots w_J | O(\gamma_t))}\\
\intertext{This ``bag of words'' assumption is not strictly true but simplifies training and inference.  The denominator can be rewritten without $O(\gamma_t)$, since the two terms are independent when $\phi$ is not known:}
p(\phi_l| w_1 \dots w_J, O(\gamma_t)) &= \frac{\prod_j p(w_j | \phi_l, O(\gamma_t)) p(\phi_l | O(\gamma_t))} {p(w_1 \dots w_J)}
\end{align}
We assume the priors are constant and, therefore, do not consider them
in the inference process.  For brevity, we drop
$\phi_l$; it is implicitly assumed $True$. 

We estimate the distribution $p(w_j | \phi_l, O(\gamma_t))$ as a
multinomial distribution using Naive Bayes.  First we rewrite with
Bayes' rule:
\begin{align}
p(w_j | O(\gamma_t)) &= \frac{p(O(\gamma_t) | w_j ) \times
 p(w_j)}{p(O(\gamma_t) | w_j ) p(w_j) + p(O(\gamma_t) | \neg w_j ) p(\neg w_j)}\\ 
\intertext{Next, we make the assumption that $O(\gamma_t)$ consists of labels, $o_1 \dots o_K$, which are independent: } 
p(w_j | o_1 \dots o_K) &= \frac{\prod_k p(o_k | w_j ) \times p(w_j )}{\prod_k p(o_k | w_j ) p(w_j ) + \prod_k p(o_k | \neg w_j ) p(\neg w)}\label{eq:naive_bayes}
\end{align}

We compute the set of labels using the objects visible at the end of
the trajectory $\gamma_t$.  We found that most of our directions
referred to landmarks that were visible at the end of the trajectory.
Even for phrases such as ``Go through the set of double doors,'' where
the doors are located in the middle of the trajectory, they are
visible from the end, so using this set of objects works well in
practice. However, this assumption may be violated for much longer
directions.

We estimate these probabilities using co-occurrence statistics from
tags from over a million images downloaded from the Flickr
website~\citep{kollar09}. For example, using this corpus, the system
can infer which bedroom is ``the baby's bedroom'' without an explicit
label, since only that room contains a crib and a changing table.
This data allows the robot to interpret language about landmarks that
may not be in its semantic map by connecting them through
co-occurrence to landmarks that are present in the semantic map.
Specifically we compute the following estimates, where $count$ is
defined as the number of captions that contain word $w$ in the Flickr
corpus:
\begin{align}
p(o_k | w_j) = \frac{count(o_k, w_j)}{count(w_j)}
\end{align}
\noindent And without $w_j$:
\begin{align}
p(o_k | \neg w_j) = \frac{count(o_k, \neg w_j)}{count(\neg w_j)}
\end{align}

We refer to the above method as the Naive Bayes model for landmark
factors.  Using this approximation leads to problems because
many words in the context $o_1 \dots o_K$ are not relevant to a
particular landmark phrase.  For example, if the robot is going to
``the kitchen,'' observations of the refrigerator and microwave will
help it identify a promising candidate region, but observations of a
door will not give much additional information.  Incorporating these
objects into the approximation in Equation~\ref{eq:naive_bayes} causes
it to underestimate the true probability of encountering a landmark
phrase.  To compensate, we use the subset of $o_1 \dots o_k$ that has
the highest probability when estimating $p(\phi_l | w_1 \dots w_J, o_1
\dots o_K)$:
\begin{align}
p(\phi_l | \lambda_l, \gamma_t) \approx \max_{O \in powerset(\{o_1 \dots o_K\})} p(\phi_l | w_1 \dots w_J, O).\label{eq:min_entropy}
\end{align}

We estimate this probability over the powerset of all objects observed
at a particular location.  This computation is tractable because of
the limited numbers of object types available in the semantic map;
typically no more than ten different types of objects will be visible
at a particular location.  We refer to this version as the salient
object model because it finds the subset $O$ that provides the
strongest evidence for $\phi_l$.  We are able to compute this term
exactly because at any particular location, relatively few unique
objects are visible, so the cardinality of the powerset of visible
tags is small.

\subsection{Inference}

Given the graphical model for the natural language command, the
inference algorithm for following directions through completely known
maps (\emph{global} inference) searches through all possible paths in
a topological map of the environment to find the maximum of the
distribution in Equation~\ref{eq:g3_argmax}.  Global inference is
performed using a dynamic programming algorithm~\citep{viterbi67} that
finds the most probable trajectory corresponding to a set of natural
language directions. The algorithm takes as input a set of starting
locations, a map of the environment with some labeled objects as well
as a topology, and the graphical model created from the parsed
directions.  It outputs a sequence of nodes in the topological map,
corresponding to trajectory through the environment.
\begin{figure}[t]
\centering
\subfigure[t=0]{
\fbox{
\includegraphics[width=0.29\linewidth]{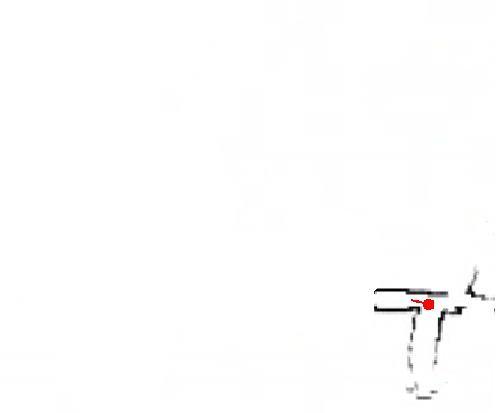}
}
}\subfigure[t=50]{
\fbox{
\includegraphics[width=0.29\linewidth]{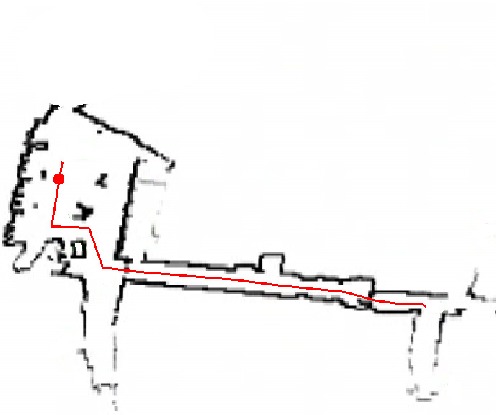}}
}\subfigure[t=100]{
\fbox{
\includegraphics[width=0.29\linewidth]{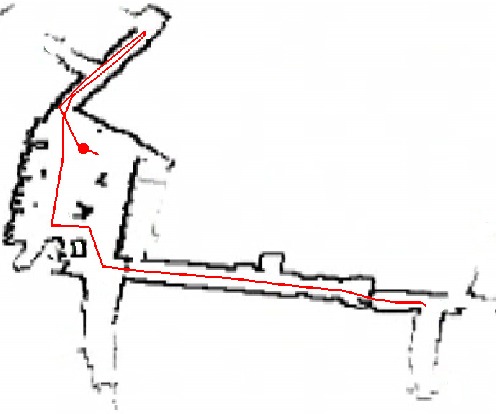}}
}
\subfigure[Command]{\fbox{\parbox{1\linewidth}{ Go through the double doors
    and past the lobby.  Go into the lounge with some couches.  Enjoy
    the view over there.  Go past the spiral staircase.  Continue
    towards the hall with some cubby holes but don't go down the
    hallway.  Instead take a right into the kitchen.  }}}

\caption{Explored map at three different phases the robot explores the
  environment.  Black shows obstacles/walls visible explored by the
  robot.  The red circle shows the robot's current location, and the
  red line shows its trajectory.  The system misunderstands the
  negation in the phrase ``don't go down the hallway,'' but backtracks
  when it does not find a kitchen.  After backtracking, it reaches the
  correct destination.  The robot's current location and past
  trajectory are shown in red.  \label{fig:map}}
\end{figure}


Following route directions with a complete map is useful for scenarios
where a robot might be able to perform both exploration and mapping
beforehand, such as in a home or building.  For less structured
environments or new environments where exploration and mapping cannot
be performed beforehand, a robot must follow commands with incomplete
information.  In this more general scenario, the robot must follow
directions without a pre-built map, discover the environment as it
navigates and update its plans when it discovers new information.  We
present two different approaches to address the scenario where the map
is not known beforehand.  The first follows the directions step by
step, exploring the map as it goes.  At each step, it chooses the next
action based on the best available action at its current location.  If
it makes a wrong turn, it does not backtrack or explore.  When the
robot reaches a region not in the existing map, it explores this
region, incrementally growing its map of the environment. We refer to
this algorithm as {\em greedy local inference}.  The second algorithm
is similar, but if there is no transition at the current location with
probability above a threshold, it backtracks and explores a different
region.  We refer to this algorithm as {\em exploring local
  inference}.  Figure~\ref{fig:map} shows the explored map at two
different phases of the exploring local inference algorithm.  We
expect global inference to perform better because it searches through
all possible paths to find the one that best matches the descriptions.
However, the local inference algorithms are more practical because
they do not need a complete map of the environment to follow
directions.



The system creates a topological roadmap from the gridmap of the
environment, then searches for a path within this graph.  The roadmap
is created by segmenting spaces based on visibility and detected
objects and then extracting a topology of the environment from this
segmentation, building on techniques described by
\citet{brunskill:tmu}.  As a robot path extends through each of the
nodes, it may take on any of the four cardinal directions, which leads
to connections in the topological map that include the Cartesian
product of the original topological map connections and the four
cardinal directions.  This enables the system to use features of the
orientation of the robot along a path to differentiate the next
correct action.  For example, ``turn right'' might be differentiated
from ``go straight'' only by the fact that the orientation at the end
of the path differs by approximately $90^\circ$ across the two choices.  For the
wheelchair domain, we assume that the robot moves in two
dimensions. For the MAV, we create a two-level topological map by
duplicating all nodes at each level, enabling it to fly through a
three-dimensional space.

\subsection{Evaluation}

To evaluate our system, we collected corpora for two application
domains: the robotic wheelchair, and the robotic MAV.  We also
demonstrated the system end-to-end on these two robotic platforms.

\subsubsection{Corpus-Based Evaluation}

For the wheelchair, we collected a corpus of route directions between
two locations in an indoor office environment.  Subjects wrote
directions as if they were directing another person through the space.
We collected directions through two environments.  Environment 1 is a
work area with a computer lab and offices ($\SI{133 x 137}{m}$), while
Environment 2 is an atrium with a cafe and classrooms (\SI{99 x
  62}{m}). We asked fifteen subjects to write directions between 10
different starting and ending locations in environment 1, and another
fifteen to write directions between 10 different location pairs in
environment 2, for a total of 300 directions.  The corpus from
Environment 1 consisted of 8799 words and 939 unique words; the corpus
from Environment 2 consisted of 6467 words and 921 unique words.

Experimenters did not refer to any of the areas by name, but instead
used codes labeled on a map. Subjects were from the general population
of MIT, between the ages of 18 and 30 years old, were proficient in
English, were unfamiliar with the test environment, and were
approximately of equal gender (47\% female and 53\% male subjects).
Sample commands from the corpus appear in
Figure~\ref{fig:exampleWheelchair}.

For the MAV domain, users were familiarized with the test environment
(which was the same as Environment 2) and were asked to instruct the
pilot to take video of seven objects in the environment, each starting
from a different location.  Objects to be inspected were difficult to
see closely from the ground and included a wireless router mounted
high on the wall, a hanging sculpture, and an elevated window.
Subjects were told the vehicle's initial pose and were asked to write
down instructions for a human pilot to fly a MAV to the desired object
and take video of that object.  The corpus consists of forty-nine
natural language commands, for a total of 2576 words and 467 unique
words.  Subjects were engineering undergraduates unfamiliar with the
system. Figure~\ref{fig:exampleMav} shows an example set of directions
from this corpus.

We evaluate the global and local inference methods, as well as several
baseline methods.  Our first baseline is human performance at
following directions.  To measure human performance, we had one person
follow each of the directions in this corpus, asking them to notify
the experimenter if they became lost or confused. In this situation,
we found that human performance at following directions in this
dataset was 85\% and 86\% for the two datasets (e.g., 15\% of the
directions were not followed correctly by people).  Qualitatively, we
observed that people would sometimes give incorrect directions (e.g.,
they would say ``right'' when they meant ``left''), or that their
spatial ability in following directions appeared to be poor.  We did
not assess human performance for the MAV domain because of the
difficulty of having untrained users follow an aerial trajectory. We
implemented two computational baselines.  The first, \emph{Random}, is
the distance between the true destination and a randomly selected
viewpoint in the map.  This estimates the complexity of the
environment without taking into account the language input and
provides a lower bound on performance.  The second, \emph{Last
  Phrase}, uses the location that best matches the last phrase in the
directions according to our model, ignoring the rest of the
directions.  This provides a lower bound that shows whether the model
is able to leverage information from the entire natural language
command.  We present all baselines using the Naive Bayes model for the
landmark factor (Equation~\ref{eq:naive_bayes}), as well as the
salient object model (Equation~\ref{eq:min_entropy}).

Table~\ref{tab:route_directions_wheelchair} shows results for the
wheelchair.  We present the fraction of commands successfully followed
to within 10 meters of the true destination.  Our global inference
model performed the best because it has complete access to the map,
demonstrating the capability of the model with complete
information. The fact that \G3 (Global Inference) outperformed the
\emph{Last Phrase} baseline demonstrates that our method is
successfully applying information from the whole set of natural
language directions, rather than just the last phrase.  We found that
performance was better in Environment 1 than in Environment 2 because
the former has a simpler topology and more distinct landmarks.  We
also see that the salient object model for the landmark factor
significantly outperformed the Naive Bayes model.  The salient object
model provides a filter that removes irrelevant terms from the Naive
Bayes approximation, so that the algorithm only uses terms known to be
relevant to the landmark phrase given by the language.  To measure the
effect of parsing on our results, we tested the salient object model
using ground truth parses.  Because of the fixed, simple parse
structure, the parsing accuracy did not matter as much, and we
observed a small effect on overall performance when using ground truth
parses.

The greedy local inference method performed very poorly; in contrast,
the exploring local inference method achieved performance competitive
with the global inference algorithm.
Table~\ref{tab:exploration_wheelchair} shows the fraction of the
environment explored by the local inference algorithms.  We report the
fraction of topological map nodes visited by the algorithm, excluding
nodes on the shortest path between the start of the trajectory and the
end.  This metric represents extra exploration by the algorithm as it
followed the directions.  The greedy local inference algorithm
explored a small part of the environment, but also achieved low
performance compared to the global inference algorithm, which had
access to the entire map.  In contrast, the exploring local inference
algorithm achieved performance competitive with global inference
without visiting the entire map first.

Table~\ref{tab:route_directions_mav} shows results for the MAV domain.
We see the same high-level patterns as in the wheelchair domain: the
salient object model outperformed the Naive Bayes model, and global
inference outperformed greedy local inference, while exploring local
inference was competitive with global inference.
Table~\ref{tab:exploration_mav} shows the fraction of explored
environment by the local inference algorithms. This consistent pattern
suggests our results will likely generalize to different domains.  In
addition, we compared the performance of the two dimensional model
(identical to the wheelchair domain) to a three dimensional model,
where the robot must infer a three-dimensional trajectory through the
environment, including heights.  We found that the 3D model
outperformed the 2D model, suggesting that the 3D model better matches
language produced in this domain, quantitatively demonstrating that
the 3D structure improves performance.

\begin{table}[t]
\begin{center}
{\small
\begin{tabular}{lrr}
\toprule
&                                                Environment 1& Environment 2\\
                                                     & \% correct & \% correct\\
\midrule
Human Performance                                    & 85\%       &  86\%    \\
Random                                               & 0\%        &   0\%    \\
\midrule
\multicolumn{3}{c}{Naive Bayes (Equation~\ref{eq:naive_bayes})}    \\
Last Phrase only                                     & 40\%       & 25\%     \\
\G3 (greedy local inference)                       & 30\%       & 20\%     \\
\G3 (exploring local inference)                    & 39\%       & 27\%     \\
\G3 (global inference)                             & 39\%       & 27\%     \\
\midrule
\multicolumn{3}{c}{Salient Objects (Equation~\ref{eq:min_entropy})}\\
Last Phrase only                                     & 50\%       & 33\%     \\
\G3 (greedy local inference)                       & 30\%       & 20\%     \\
\G3 (exploring local inference)                    & 71\%       & 54\%     \\
\G3 (global inference)                             & 71\%       & 57\%     \\
\G3 (global inference, annotated parses)           & 67\%       & 59\%     \\
\bottomrule
\end{tabular}
}
\end{center}
\caption{Performance at following directions to within 10 meters of
  the true destination in our two test environments, for directions
  given to a robotic wheelchair.\label{tab:route_directions_wheelchair}}
\end{table}

\begin{table}{t}
\begin{center}
{\small
\begin{tabular}{lrr}
\toprule
&                                                Environment 1& Environment 2\\
\midrule
\multicolumn{3}{c}{Naive Bayes (Equation~\ref{eq:naive_bayes})}       \\
\G3 (greedy local inference)                    &  2\%      & 2\%   \\
\G3 (exploring local inference)                 &  62\%     & 61\%  \\
\midrule
\multicolumn{3}{c}{Salient Objects (Equation~\ref{eq:min_entropy})}\\
\G3 (greedy local inference)                    &   1\%     &  2\%  \\
\G3 (exploring local inference)                 &   77\%    & 70\%  \\
\bottomrule
\end{tabular}
}
\end{center}
\caption{Fraction of the environment explored for each
  algorithm.\label{tab:exploration_wheelchair}}
\end{table}

\begin{table}[t]
\begin{center}
{\small
\begin{tabular}{lr}
\toprule
                                   & Environment 2\\
                                   & \% correct   \\
\midrule
Human Performance                  & --           \\
Random                             & 0\%          \\
\midrule
\multicolumn{2}{c}{Naive Bayes (Equation~\ref{eq:naive_bayes})} \\
Last Phrase only                   & 20\%         \\
\G3 (greedy local inference)     & 0\%          \\
\G3 (exploring local inference)  & 20\%         \\
\G3 (global inference)           & 22\%         \\
\midrule
\multicolumn{2}{c}{Salient Objects (Equation~\ref{eq:min_entropy})}\\
Last Phrase only                   & 33\%        \\
\G3 (greedy local inference)     & 0\%         \\
\G3 (exploring local inference)  & 55\%        \\
\G3 (global inference, 2D)       & 55\%        \\
\G3 (global inference, 3D)       & 65\%        \\
\bottomrule
\end{tabular}
}
\end{center}
\caption{Performance at following directions to within 10 meters of
  the true destination in our environments for directions given to a
  robotic MAV.\label{tab:route_directions_mav}}
\end{table}

\begin{table}[t]
\begin{center}
{\small
\begin{tabular}{lrr}
\toprule
                                   & Environment 2\\
                                   & \% explored\\
\midrule
\multicolumn{2}{c}{Naive Bayes (Equation~\ref{eq:naive_bayes})} \\
\G3 (greedy local inference)     &    1\%      \\
\G3 (exploring local inference)  &    49\%     \\
\midrule
\multicolumn{2}{c}{Salient Objects (Equation~\ref{eq:min_entropy})}\\
\G3 (greedy local inference)     &      1\%     \\
\G3 (exploring local inference)  &     49\%     \\
\bottomrule
\end{tabular}
}
\end{center}
\caption{Performance at following directions to within 10 meters of
  the true destination in our environments for directions given to a
  robotic MAV.\label{tab:exploration_mav}}
\end{table}

\subsubsection{Real-world Demonstration}
\begin{figure}[t]
\centering
\includegraphics[width=0.32\linewidth]{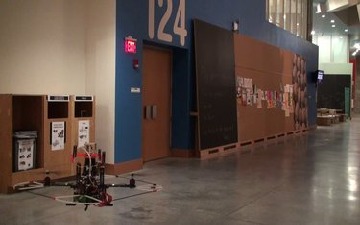}\hfil%
\includegraphics[width=0.32\linewidth]{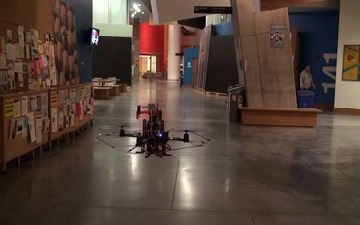}\hfil%
\includegraphics[width=0.32\linewidth]{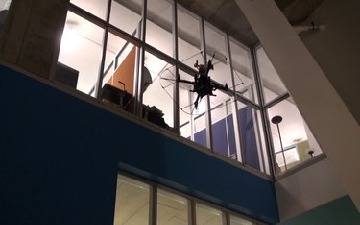}\\
\includegraphics[width=0.32\linewidth]{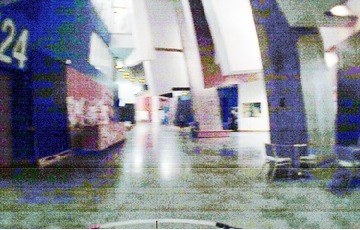}\hfil%
\includegraphics[width=0.32\linewidth]{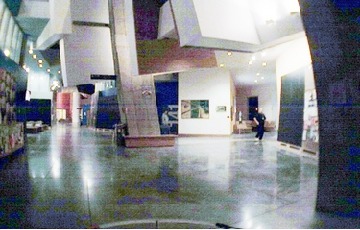}\hfil%
\includegraphics[width=0.32\linewidth]{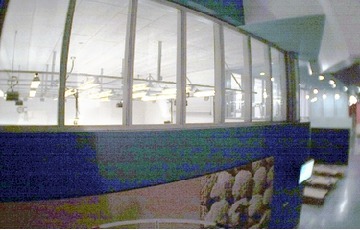}
\caption{(top) Photographs of the MAV executing an interactive
  series of instructions.  (bottom) Imagery from the on-board camera,
  transmitted to the operator as the MAV flies.}
\label{fig:mav_interactive}
\end{figure}
\begin{figure}[t]
\centering
\includegraphics[width=0.32\linewidth]{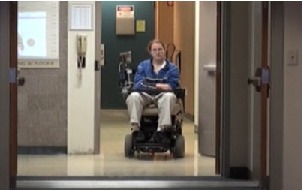}\hfil%
\includegraphics[width=0.32\linewidth]{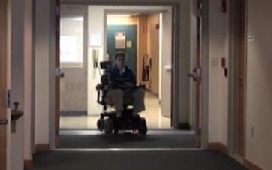}\hfil%
\includegraphics[width=0.32\linewidth]{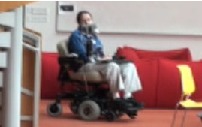}
\caption{Photographs of the wheelchair following the command, ``Go
  through the double doors, past the lobby.  Go into the lounge with
  some couches. Enjoy the view.  Go past the spiral staircase.
  Continue towards the cubby holes.  Don't go down the hallway.
  Instead take a right into the kitchen.''}
\label{fig:wheelchair_interactive}
\end{figure}
We evaluated the \G3 framework end-to-end at following directions on
the wheelchair (moving in two dimensions) and a robotic MAV (moving in
three dimensions).  Our autonomous wheelchair, shown in
Figure~\ref{fig:wheelchair}, is equipped with laser range scanners for
obstacle sensing, navigation, and localization.  We initialized it
with a semantic map of the environment, in which we labeled the
locations of known objects.  Given a typed or spoken command, it
inferred a trajectory through the environment using the \G3
framework. Once the trajectory was inferred, the vehicle executed the
trajectory autonomously.  Figure~\ref{fig:wheelchair_interactive}
shows photos of the wheelchair as it follows natural language
commands; see \url{http://youtu.be/yLkjM7rYtW8} for video.

Our MAV, shown in Figure~\ref{fig:mav}, is the AscTec Pelican
quad-rotor helicopter, manufactured by Ascending Technologies GmBH.
We outfitted the vehicle with both LIDAR and camera sensors, which
allows us to obtain accurate information about the environment around
the vehicle.  In previous work~\citep{bachrachEMAV09} we developed a
suite of sensing and control algorithms that enable the vehicle to
explore unstructured and unknown GPS-denied environments. Here, we
leverage that system to localize and control the vehicle in a
previously explored, known environment~\citep{He08ICRA,GrzonkaICRA09}.
We developed an interface that enabled a person to give directions to
the MAV using a speech recognition system or by typing a textual
string.  Paths computed by the \G3 framework were then executed
autonomously by the MAV. Figure~\ref{fig:mav_interactive} shows scenes
from the MAV following a set of directions; see
\url{http://youtu.be/7nUq28utuGM} for video of the
end-to-end system.  Directions that our method handled 
successfully on the robotic platform include:
\begin{itemize}
\item [(a)] ``Go past the library and tables till you see a cafe to the left. Fly
past the cafe and there will be other eateries.  Head into this area.''
\item [(b)] ``Stand with your back to the exit doors.  Pass the cafe on your
right.  Make a right directly after the cafe, and into a seating area.  Go
towards the big question mark.''
\item [(c)] ``Go straight away from the door that says CSAIL, passing a room on
your right with doors saying MIT Libraries.  Turn left, going around the cafe
and walk towards the cow.''
\item [(d)] ``Turn right and fly past the libraries.  Keep going straight and on the
left near the end of the hallway there is a set of doors that say Children's
technology center.  You are at the destination.''
\item [(e)] ``Fly to the windows and then go up.''
\end{itemize}

Our robotic experiments demonstrate that our approach is practical for
realistic scenarios.  The \G3 framework can quickly and robustly infer
actions to take in response to natural language commands.

\section{Discussion}

\begin{figure}
\centering
\subfigure[Mobile manipulation.]{\includegraphics[width=0.3\linewidth]{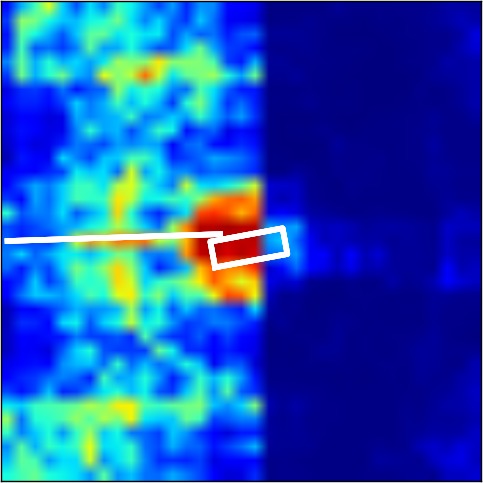}}%
\subfigure[Route directions.]{\includegraphics[width=0.3\linewidth]{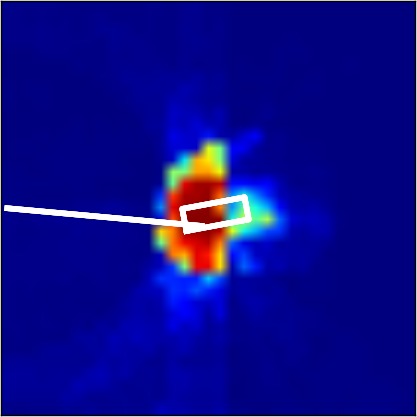}}
\caption{Heat map generated for the phrase ``to the truck'' starting at the left side of the scene.  The truck is indicated with the white rectangle.\label{fig:heatmap}}
\end{figure}

Our approach is a significant step forward in terms of advancing
robotic language understanding, however significant steps forward
remain.  As previously noted, our approach enables the robot to learn
probabilistic predicates for words such as ``to'' and ``near.''  It is
instructive to explore qualitatively what is learned.
Figure~\ref{fig:heatmap} shows a heatmap generated by querying the
mobile manipulation and route direction model for the phrase ``to the
truck.''  Each cell in the heat map is generated by assessing the
value of the $p(\phi | \mbox{``to''}, \gamma_{truck},
\gamma_{path},M)$ for a straight line path that starts at the left
side of the heat map and goes to that pixel.  Red corresponds to high
probability and blue corresponds to low probability.  The highest
probability path is drawn in white, and the truck's geometry (viewed
from the top) is the white rectangle.  Both models learn reasonable
models for ``to''; the hot spots are close to the landmark of the
truck.  Note that these probabilistic predicates were learned
automatically by the system from lower-level features without
providing a geometric definition for the word ``to.''

A significant limitation to our approach is that it must be adapted to
the domain it is being used on.  At minimum this adaptation requires
training data including language and groundings from the specific
domain.  Sometimes additional changes might be necessary, such as a
parser that is capable of handling language in the specific domain, as
well as specific features that capture the semantics of words in the
domain.  Features are currently constructed manually; with new
advances in deep learning it is intriguing to consider automatic
feature construction.  However a key question is to define the input
representation - if the robot has already detected and localized
objects, then the deep learning representation must learn
probabilistic predicates over a different space from the low-level
image space.

Our approach could be extended to other languages given data, but it
requires adding additional features specific to those languages.  For
example, Korean distinguishes between two forms of the English
preposition ``in,'' one for when the objects are tight-fitting, and a
second when they are loose fitting~\citep{norbury08}.  Our approach
would need additional features to enable the framework to learn this
concept.

An additional limitation is the parsers that we are building on.  Our
approach used existing technology to parse the sentences.  Using
parsers trained on external datasets has potential because it can lead
to increased performance by leveraging the external datasets.  However
domain mismatches can also cause problems.  For example, the Stanford
parser often parsed phrases such as ``down the hall'' incorrectly
because the word ``down'' is most often used as an adverb in the
training set.  It would be useful to retry these experiments with more
recent parsers which might not suffer from these deficiencies and to
adapt parsers to our domains.  Another approach, which others have
taken, is to jointly train a parser along with semantic understanding
modules~\citep{kollar13a, matuszek12, artzi13}.  These approaches
require predicates to be provided by the system designer, but allow
the parser to adapt to domain specific utterances and language,
including imperative commands which are not often found in newswire
text. \citet{kollar13a} extends \citet{matuszek12joint} to include
relational predicates.

A key challenge in our real-world demonstrations was obtaining an
accurate and complete world model for the robot.  In the case of the
robotic forklift, there were many objects in the environment that were
not in the robot's world model (such as a pile of dirt and parked
cars) and thus could not be referred to using language.  This problem
is pervasive in robotics: people want to talk to the robot about
everything they can see, but the robot does not have a model of
everything the person can see, due to limitations in its sensors and
perceptual abilities.

\section{Conclusions}
\label{chap:conclusions}

In this paper, we have taken steps toward robust spatial language
understanding using the \G3 framework.  Our approach learns grounded
word meanings that map between aspects of the external world. The \G3
framework defines a probabilistic graphical model dynamically
according to the compositional, hierarchical structure of language,
enabling word meanings to be combined to understand novel commands not
present in the training set.  We have demonstrated that the \G3
framework can model word meanings for a variety of mobile-manipulator
robots, such as a forklift, the PR2, a wheelchair, and a robotic MAV.
Our more recent work has shown how the probabilistic framework
described here can be trained with less supervision~\citep{tellex13}.
We have also adapted it to ask targeted questions when the robot is
confused by measuring entropy of the marginal distributions over
specific grounding variables~\citep{deits13}. \citet{knepper13}
demonstrate how to invert the model, searching for language that
corresponds to groundings, rather than groundings that match the
language, in order to generate natural language requests for help.

The \G3 framework is a step toward robust language understanding
systems, but many challenges remain.  A key limitation is the
requirement to define the search space for values of the grounding
variables.  This search space must be manually defined and tuned,
because if it is too large, the search process is intractable, and if
it is too small, then it is impossible to understand the command.  For
example, in the mobile manipulation domain, some commands referred to
``the empty space'' in a row of pallets, referring to an object with
no physical instantiation at all.  This problem is particularly
important when understanding longer commands; each linguistic
constituent which must be grounded increases the size of the search
space during inference.  Defining the search space itself dynamically
based on the language, and searching it efficiently, remain open
problems.  Another remaining challenge is the acquisition of
generalizable world meanings.  In this paper we learned word meanings
from large datasets, but we used word meanings tailored to each
domain, and we have not demonstrated that those learned meanings
generalize to different domains.  In some cases, word meanings
dynamically create a visual classifier, as in ``Cook until the
cornbread is brown,'' which requires recognizing when an object, which
is changing over time, takes on a certain appearance. Developing a
common framework for representing word meanings and learning a large
vocabulary of words in this framework is a source of ongoing research.
Related to this issue is the training data required to learn good
models.  A fourth problem is more complex linguistic structures.  For
example, conditional statements such as ``If a truck comes into
receiving, empty all the tire pallets into storage alpha,'' requires
identifying the event of the truck's arrival before acting.  Our
long-term research program is to develop a probabilistic framework for
enabling people to flexibly interact with robots using language.

\section{Acknowledgments}
This work was sponsored by the Robotics Consortium of the U.S Army
Research Laboratory under the Collaborative Technology Alliance
Program, Cooperative Agreement W911NF-10-2-0016, and by the Office of
Naval Research under MURI N00014-07-1-0749.  We thank
Chad Jenkins and the Brown Robotics lab for their help with the PR2
demonstration.

\bibliography{main}
\bibliographystyle{apalike}

\appendix
\section{Appendix:  Features}
\label{sec:appendix}

We present a representative set of features used in our work below.
Features not included here include distance and contact-based features
intended to capture the semantics of words such as ``on'' and
``near.''  Many spatial prepositions can be defined in terms of the
coordinate axes of the landmark object~\citep{talmy05, landau93}.  For
example, ``across'' requires the figure to be perpendicular to the
major axis of the landmark~\citep{talmy05,landau93}: to cross a road,
one most go from one side to the other, and not from one end to the
other.  In order to define features that capture the meaning of words
such as ``across,'' we must define an algorithm for finding the major
axis of the landmark. In many contexts, there is no single set of
axes: for example, there are many paths across a square room.  We
solve this problem by computing the unique axes that the figure
imposes on the landmark and then quantifying how well those axes match
the landmark.  The system computes these axes by finding the line that
connects the first and last point in the figure and extending this
line until it intersects the landmark.  The origin of the axes is the
midpoint of this line segment, and the endpoints are the two points
where the axes intersect the landmark.

\citet{talmy05} defines ``past'' as figure's path must be going
perpendicular to a point $P$ ``at a proximal remove'' from the
landmark.  The path of the figure must be perpendicular to a line
going from the landmark to this point.  We define the ``past axes'' as
this line and a line perpendicular to it.

\begin{figure}[t]
\centering
\includegraphics[width=0.5\linewidth]{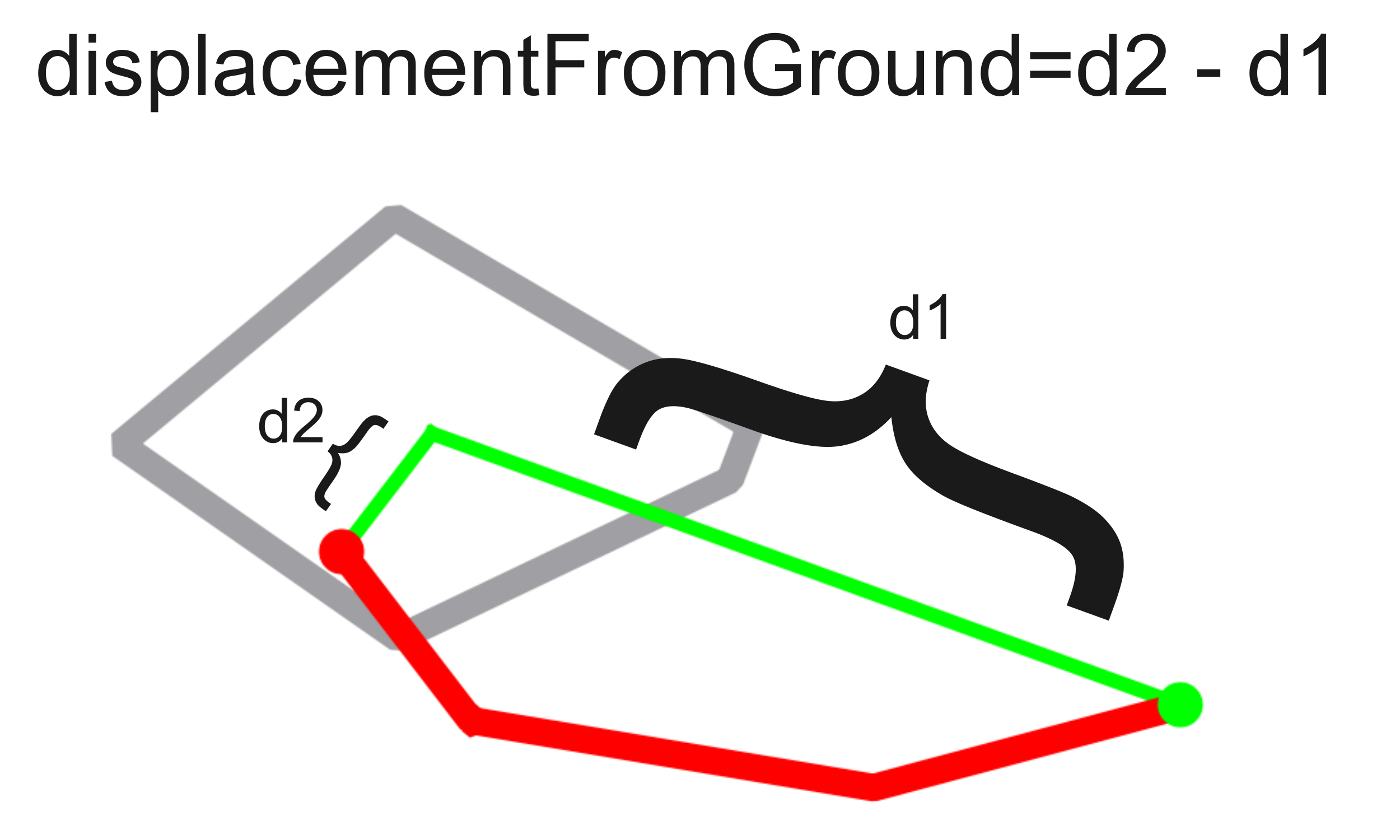}
\caption{Illustration of the computation of the feature {\em
    displacementFromLandmark}.\label{fig:displacement}}
\end{figure}

\begin{itemize}
\item {\bf angleBtwnLinearizedObjects}: The angular difference of the
  slope of lines fit to the points in the figure and points in the
  boundary of the landmark, using linear regression.
\item {\bf angleFigureToPastAxes}: The angle between a line fit to
  points in the figure using linear regression and the line between
  the figure and the landmark at the closest point.
\item {\bf averageDistStartEndLandmarkBoundary}: The average of the
  distance between the start of the figure and the boundary of the
  landmark, and the distance between the end of the figure and the
  boundary of the landmark.
\item {\bf displacementFromLandmark}: The difference in distance
  between the start point of the figure to the landmark, and the end
  point of the figure to the landmark, illustrated in
  Figure~\ref{fig:displacement}.
\item {\bf distAlongLandmarkBtwnAxes}: The distance along the
  perimeter of landmark between the start and end of the minor axis.
\item {\bf distStartLandmarkBoundary}: The distance of the start of
  the figure to the boundary of the landmark.
\item {\bf distFigureEndToLandmark}: The distance from the end point
  of the figure to the boundary of the landmark.
\item {\bf distFigureStartToLandmark}: The distance from the start
  point of the figure to the landmark.
\item {\bf eigenAxesRatio}: The ratio between the eigenvectors of the
  covariance matrix of the landmark when represented as an occupancy
  grid.
\item {\bf figureCenterOfMassToAxesOrigin}: The distance between the
  center of mass of the points in the figure and the axes origin.
\item {\bf figureCenterOfMassToLandmarkCentroid}: The distance between
  the center of mass of the figure and the centroid of the landmark.
\item {\bf pastAxesLength}: The distance between the figure and the
  landmark, at the point the figure most closely approaches the
  landmark.
\item {\bf peakDistToAxes}: The maximum distance between the figure
  and the axes it imposes on the landmark, for the part of the figure
  that is inside the landmark.
\item {\bf ratioFigureToAxes}: The ratio between the distance between
  the start and end points of the figure and the length of the major
  axis.
\item {\bf stdDevToAxes}: The standard deviation of the distances
  between the figure and the major axis of the landmark, stepping
  along the figure.
\end{itemize}

Many of these features are based on distances.  However, spatial
relations are scale invariant: one can use a word like ``across'' to
describe a small-scale event such as crab crawling across a rock, and
a large-scale event such as a car driving across the country.
Distance features such as {\em distFigureEndToLandmark} and {\em
  displacementFromLandmark} are made scale invariant by normalizing by
the bounding box of the scene.  The normalization prevents learned
models from overfitting to a particular geometric scale.

\subsection{Verb Features for Route Directions}

Each type of directive corresponds to an expected turn amount,
$\theta_{\lambda_v}$, which is $0^\circ$ for ``straight,'' $-90^\circ$
for ``right,'' and $90^\circ$ for ``left.''  Next, we define
$\theta_{\gamma_t}$, the actual amount that a robot would turn when
following the trajectory.  Finally, we use a sigmoid distribution on
the difference between the expected turn amount and actual turn
amount:
\begin{align}
   p(\phi_v | \lambda_v, \gamma_t) \approx \frac{1}{1 - e^{|\theta_{\lambda_v} - \theta_{\gamma_t}|}}
\end{align}

In aerial task domains, commands also include three-dimensional
components, such as ``fly up'' or ``fly down.''  If $\lambda_v$
contains one of these directives, we check whether the trajectory ends
at a higher or lower elevation than the start, or stays the same.  If
the trajectory's elevation matches the directive, we set $p(\phi_v |
\lambda_v, \gamma_t) = 1$, and otherwise, $0.000001$.

\end{document}